%% file: neurips_main.tex
\documentclass{article}

 \usepackage[preprint]{neurips_2025}

\usepackage[utf8]{inputenc} 
\usepackage[T1]{fontenc}    
\usepackage{hyperref}       
\usepackage{url}            
\usepackage{booktabs}       
\usepackage{amsfonts}       
\usepackage{nicefrac}       
\usepackage{microtype}      
\usepackage{xcolor}         
\definecolor{citeblue}{RGB}{0,114,178} 
\hypersetup{
    colorlinks=true,
    citecolor=citeblue,
    linkcolor=black,
    urlcolor=citeblue
}

\usepackage{microtype}
\usepackage{graphicx}
\usepackage{subcaption}
\usepackage{bbm}
\usepackage{pgfplotstable}
\usepackage{siunitx}

\usepackage{amsmath}
\usepackage{amssymb}
\usepackage{mathtools}
\usepackage{amsthm}
\usepackage{enumitem}

\input{math_commands.tex}

\usepackage[capitalize,noabbrev]{cleveref}
\usepackage{placeins}
\usepackage{float}
\title{Correcting Influence:
Unboxing LLM Outputs with Orthogonal Latent Spaces}

\author{%
  Shixing Yu \\
  Electrical and Computer Engineering\\
  Cornell Tech\\
  \texttt{sy774@cornell.edu} \\
  \And
  Promit Ghosal \\
  Department of Statistics \\
  University of Chicago\\
  \texttt{promit@uchicago.edu} \\
  \And
  Kyra Gan \\
  Operations Research and Industrial Engineering\\
  Cornell Tech\\
  \texttt{kyragan@cornell.edu} \\
}

\begin{document}

\maketitle
\input{_s0_abs}
\input{_s1_intro}

\input{_s2_related_work}
\input{_s3_prelim}

\input{_s4_method}
\input{_s5_experiment}

\input{_s6_conclusion}
\input{_s7_acknowledgement}

\bibliography{example_paper}
\bibliographystyle{plainnat}

\FloatBarrier
\clearpage
\appendix
\counterwithin{table}{section}
\counterwithin{figure}{section}
\section*{Appendix}

\input{_sa_appendix.tex}

\end{document}

%% file: math_commands.tex
\usepackage{amsmath,amsfonts,bm, amsthm, mathtools}
\theoremstyle{plain}
\newtheorem{theorem}{Theorem}[section]

\theoremstyle{definition}
\newtheorem{definition}[theorem]{Definition}

\theoremstyle{remark}
\newtheorem{remark}[theorem]{Remark}

\newcommand{\ttrain}{\text{train}}
\newcommand{\ttest}{\text{test}}
\newcommand{\enc}{\text{enc}}
\newcommand{\dec}{\text{dec}}
\newcommand{\pre}{\text{pre}}

\def\eqref#1{Eq.~(\ref{#1})}

\def\1{\bm{1}}

\DeclareMathAlphabet{\mathsfit}{\encodingdefault}{\sfdefault}{m}{sl}
\SetMathAlphabet{\mathsfit}{bold}{\encodingdefault}{\sfdefault}{bx}{n}

\newcommand{\R}{\mathbb{R}}



%% file: _s0_abs.tex
\begin{abstract}
A critical step for reliable large language models (LLMs) use in healthcare is to attribute predictions to their training data, akin to a medical case study.
This requires token-level precision: pinpointing not just which training examples influence a decision, but which 
tokens within them are responsible.
While \emph{influence functions} 
offer a principled framework for this, prior work is restricted to \emph{autoregressive} settings and relies on an implicit assumption of \emph{token independence}, rendering their identified influences unreliable.
We introduce a flexible framework that infers \emph{token-level influence} through a latent mediation approach for \emph{general prediction tasks}. Our method attaches 
\emph{sparse autoencoders}  to any layer of a pretrained LLM to learn a basis of approximately independent latent features.
Unlike prior methods where influence decomposes additively across tokens, 
influence computed over latent features is inherently \emph{non-decomposable}.
To address this, we introduce a novel method  using \emph{Jacobian-vector products}.
Token-level influence is obtained by propagating latent attributions back to the input space via token activation patterns.
We scale our approach using efficient inverse-Hessian approximations.
Experiments on medical benchmarks show our approach identifies sparse, interpretable sets of tokens that \emph{jointly} influence predictions.
Our framework enhances trust and enables model auditing, generalizing to any high-stakes domain requiring transparent and accountable decisions.
\end{abstract}

%% file: _s1_intro.tex
\vspace{-5pt}
\section{Introduction}
\vspace{-5pt}
The deployment of LLMs in high-stakes domains like healthcare hinges on a critical and unmet requirement: the ability to audit a model's reasoning by tracing its predictions directly to the evidence in its training data. 
This need for verifiability is urgent, as LLMs are increasingly explored for clinical tasks such as diagnostic support and treatment planning, where errors can have severe consequences \citep{singhal2023large, topol2019high}. Without this capability—akin to a clinician demanding the source for a medical decision—LLMs remain unverifiable black boxes. Their tendency to hallucinate \citep{ji2023survey} and their susceptibility to spurious correlations present in training data \citep{oberst2019counterfactual} pose significant safety risks, undermining the trust required for clinical adoption \citep{futoma2020myth, ghassemi2021false}.

This fundamental need for evidence-based reasoning is not adequately addressed by prevailing interpretability methods. Techniques like Chain-of-Thought 
prompting generate rationales that are often post hoc justifications rather than faithful reflections of the model's true decision process \citep{turpin2023language, barez2025chain}. Other popular approaches, such as attention visualization \citep{wiegreffe2019attention, jain2019attention} or gradient-based feature attribution \citep{sundararajan2017axiomatic}, are limited to explaining a single forward pass of a model. They operate within the context of a given input, providing no insight into how prior training experiences shaped the model's fundamental behavioral patterns and knowledge \citep{feldman2020neural}.
This represents a critical limitation for clinical deployment, where the ability to pinpoint the exact training evidence behind a prediction---not just generate plausible-sounding rationales---is essential for medical professionals to validate the model's logic against established knowledge, fact-check its conclusions, and ultimately build the trust required for adoption in safety-critical settings.

A principled framework for addressing this question lies in \emph{influence functions} (IFs), a tool from robust statistics that explains
how a model's predictions depend on its training data \citep{hampel1974influence}. 
This approach 
treats the model as an empirical entity shaped by its dataset, enabling one to trace a final prediction back to influential training points \citep{koh2017understanding}. Recent work has successfully scaled this approach to modern LLMs, demonstrating its potential to reveal generalization patterns by attributing influence down to the {token level} \citep{grosse2023studying}. However, a key limitation persists: 
the IF framework assumes independence among the components of the objective (e.g., tokens in an autoregressive prediction task in prior work). This assumption is necessary for influence scores to be meaningfully interpretable, as it ensures that the relative difference in influence between components is well-defined. In practice, 
the tokens within LLMs are highly correlated. Thus, prior implementations, while powerful, produce influence estimates that are theoretically unsound and difficult to interpret 
\citep{basu2020falsifiability,tsimpoukelli2021multimodal}.

We introduce a robust framework that infers \emph{token-level} influence on test predictions via latent mediation, enabling more reliable influence estimation. Building on recent monosemanticity research \citep{bricken2023monosemanticity, templeton_scaling_2024} and disentangled representation learning \citep{wang2024disentangled}, our method leverages the fact that neural networks decompose into semantically meaningful, independent components. Our method generalizes to \emph{general prediction tasks} by propagating influence through disentangled latent spaces where features exhibit statistical independence, critical for reliable influence estimation. 
Our contributions are fourfold:
\begin{enumerate}[leftmargin=*, itemsep=0pt, topsep=0pt, partopsep=0pt, parsep=1pt]
    \item \textbf{Unified sample- and feature-level influence}: We extend influence analysis 
    beyond the isolated-token paradigm of prior work to model the \emph{joint influence of tokens} within training sample-label pairs.
    By propagating influence from latent features to input tokens through their joint activation patterns, we attribute
    predictions to specific token combinations in the training data while leveraging monosemantic structure. Unlike methods treating neurons as atomic units, we recognize meaningful computation occurs at interpretable feature level spanning multiple neurons.
    \item \textbf{Stable, independent feature extraction via sparse autoencoders (SAEs)}: We use SAEs \citep{gao2024scaling, cunningham2023sparse, marks2024sparse, cong2023sparse} as an interpretability component to produce sparse, approximately orthogonal latent features at an intermediate layer. We then compute influence scores with respect to these latent features, improving the stability and interpretability of training-data attributions.

    \item \textbf{Scalable non-decomposable influence estimation via derivative swapping}:  Latent-level influence is holistically interdependent and lacks the additive decomposition of token-level approaches. While naive Jacobian-vector product (JVP) evaluation requires an $\mathcal{O}(d_l)$ forward-mode pass per feature, we exploit Clairaut's Theorem to swap the derivative order. This gradient-derivative formulation restructures the computation into a single reverse-mode pass, reducing complexity to $\mathcal{O}(1)$ and achieving a $10\times$ to $20\times$ practical speedup. 

    \item \textbf{Large-scale empirical validation on medical and general reasoning}: We evaluate our framework on 1B and 1.5B parameter models (Llama-3.2 and Qwen2.5) across multiple QA datasets. Rigorous necessity and sufficiency ablations show our method isolates compact, highly influential circuits that systematically outperform activation magnitude and frequency baselines. 
    Moreover, heatmap visualizations over input tokens on fixed test samples suggest potential patterns in the model’s behavior, revealing that incorporating context during training leads to different behavior than not doing so; however, further investigation is needed to fully understand this behavioral difference.
\end{enumerate}

\begin{figure*}[t]
\centering
\includegraphics[width=\textwidth]{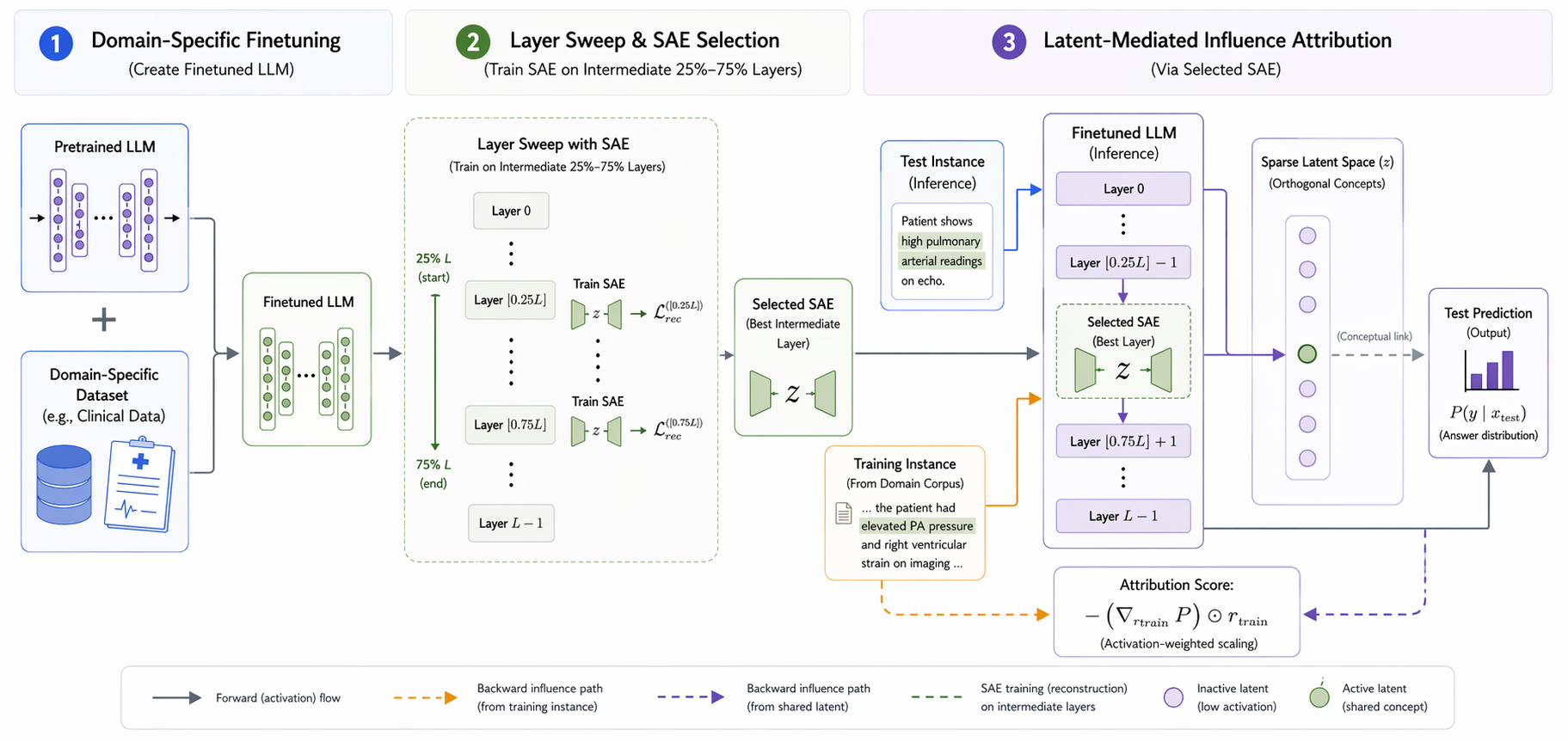}
\caption{
\textbf{Pipeline overview.} 
    Overview of RepInfLLM. A domain-specific LLM is first finetuned, then SAEs are swept over intermediate layers (25\%–75\%) to select a representative latent space. During inference, the selected SAE is inserted inline to map both training and test instances into shared sparse latents, enabling influence attribution directly in representation space. The prediction follows the standard forward pass, while attribution is computed via a backward path that operates on disentangled latents using activation-weighted scaling, avoiding token-level entanglement.
}
\label{fig:figure1}
\end{figure*}

By unifying data-level and feature-level attribution, our approach offers a principled pathway toward transparent, trustworthy, and deployable LLMs for high-stakes domains, with additional potential for large-scale training data auditing and diagnostics, which we further discuss
in Section~\ref{sec:discussion}.
Section~\ref{sec:prelim} introduces our notation and preliminaries on IF and JVP. We then describe our method in Section~\ref{sec:method} and evaluate its performance in Section~\ref{sec:exp}. Additional related works are included in Appendix \ref{sec:related_work}.
The full pipeline is demonstrated in Figure~\ref{fig:figure1}.

%% file: _s2_related_work.tex
\vspace{-5pt}
\section{Related Works}\label{sec:related_work}
\vspace{-5pt}

\noindent \textbf{Interpretability in LLM\;\;}
Interpretability methods range from black-box approaches like perturbation and sensitivity analysis \citep{casalicchio_visualizing_2018, ribeiro_why_2016, covert_explaining_2021, warstadt_blimp_2020}, to 
gradient-based attribution methods
\citep{smilkov_smoothgrad_2017, sundararajan2017axiomatic, bach_pixelwise_2015, shrikumar_learning_2017, selvaraju_gradcam_2016, bilodeau_impossibility_2024}, and concept-based representations probing \citep{belinkov_probing_2022, kornblith_similarity_2019, bansal_revisiting_2021, burns_discovering_2023, zou_representation_2023, arditi_refusal_2024}. More recent work in mechanistic interpretability focuses on reverse-engineering internal model structures through circuit analysis  \citep{olah_building_2018, elhage_mathematical_2021, elhage_toy_2022}, 
and feature discovery
\citep{bricken2023monosemanticity, sharkey_circumventing_2022, cunningham2023sparse, deng_measuring_2023}. In addition to monosemanticity and disentanglement, this line of work has enabled analyses of
motifs like induction heads or copy suppression \citep{olsson_incontext_2022, mcdougall_copy_2023, cammarata_curve_2020, cammarata_curve_2021}, universality \citep{chan_natural_2023, gurnee_finding_2023, marchetti_harmonics_2023}, and emergent world models \citep{li_emergent_2023, nanda_actually_2023, ivanitskiy_structured_2023, karvonen_emergent_2024, shanahan_role_2023, janus_simulators_2022}.
Unlike these approaches, which often prioritize global model understanding, our method emphasizes actionable, testable attributions tailored for high-stakes domains like healthcare, where rapid fact-checking and validation of model decisions are critical for reliability and trust.

\noindent \textbf{Sparse Autoencoders and Independent Features\;\;}
SAEs learn disentangled, interpretable features via sparsity constraints (e.g., L1 penalty), promoting statistical independence in latent representations. 
This approach builds upon a long history of seeking independent data components, including classical linear methods like Principal Component Analysis (PCA) \citep{jolliffe2016principal} and Independent Component Analysis (ICA) \citep{hyvarinen2013independent}, as well as nonlinear probabilistic frameworks like Variational Autoencoders (VAEs) \citep{kingma2013auto}. However, SAEs offer a uniquely transparent and deterministic pathway to feature learning that balances sparsity and reconstruction fidelity.
They are widely used for mechanistic interpretability in LLMs 
\citep{cunningham2023sparse, bricken2023monosemanticity, templeton2024scaling, marks2024sparse},
with variants including
$k$-sparse SAEs \citep{makhzani2013k}, gated and JumpReLU SAEs \citep{rajamanoharan_improving_2024}, and TopK methods
\citep{gao2024scaling, bussmann2024batchtopk}.
Beyond language, SAEs extend to multimodal domains \cite{surkov2025unpacking}, radiology and medical imaging \citep{abdulaal2024x}, and reinforcement learning alignment \citep{yin2024direct}, demonstrating versatility across tasks. 
Recent work shows that transcoders (which approximate dense MLP behavior via wider, sparsely-activating networks) often match or exceed SAEs in interpretability and fidelity~\citep{dunefsky2024transcoders}.
Extending our framework to handle independent logits from a transcoder is promising but beyond the scope of this work.

\noindent \textbf{Monosemanticity and Disentanglement}
The pursuit of monosemantic features, where neurons respond to single coherent concepts, represents a major focus in interpretability research. This effort addresses the phenomenon of polysemanticity, explained through the superposition hypothesis
\citep{olah_building_2018, elhage_mathematical_2021, elhage_privileged_2023, scherlis_polysemanticity_2023, henighan_superposition_2023}. 
Solutions include both architectural modifications such as 
$k$-sparse autoencoders \citep{makhzani2013k}, softmax linear units \citep{elhage_softmax_2022, rajamanoharan_improving_2024}, as well as post-hoc methods like SAEs 
\citep{bricken2023monosemanticity, sharkey_circumventing_2022, cunningham2023sparse, deng_measuring_2023}.
Studies have examined the linearity of representations
\citep{nanda_200algorithm_2023, engels_not_2024, omahony_disentangling_2023, hendel_incontext_2023, todd_function_2023, hernandez_linearity_2023, chanin_identifying_2023, tigges_language_2024, arditi_refusal_2024}, 
identified counterexamples such as circular features
\citep{engels_not_2024} and non-linear perspectives \citep{black_interpreting_2022}.
Geometry-aware analyses show structured organization \citep{park_geometry_2024}, and scaling studies \citep{templeton2024scaling} suggest disentanglement improves with model size.
While these works aim for complete monosemanticity, our approach uses SAEs to obtain approximately independent features specifically to enable more reliable influence estimation, prioritizing practical interpretability over full disentanglement.





%% file: _s3_prelim.tex
\section{Preliminaries}\label{sec:prelim}

Given a training dataset $\mathcal{D} = \{z_i = (x_i, y_i)\}_{i=1}^n$ i.i.d. drawn from an unknown distribution, with input $x_i \in \mathcal{X}$ and label $y_i \in \mathcal{Y}$. 
A model $h_\theta : \mathcal{X} \to \mathcal{Y}$ with parameters $\theta \in \mathbb{R}^p$ is trained by minimizing the empirical risk
    $\hat{\theta} = \arg\min_{\theta} \frac{1}{n}\sum_{i=1}^n \ell(h_\theta(x_i), y_i)$,
where $\ell(\cdot,\cdot)$ is the loss function.  

\paragraph{Influence Functions (IFs)}
In statistical estimation, the IF quantifies the sensitivity of an estimator to infinitesimal perturbations in the data, under the assumption that the data are independent.
This concept extends directly to machine learning, where the high-dimensional ``parameter'' is the set of weights $\hat{\theta}$ of a trained neural network—a complex function of the data shaped by the architecture, loss, and optimizer. Once training is complete and the model parameters $\hat{\theta}$ are fixed, we can analyze their local sensitivity to individual training samples.
This is first formalized by
the \emph{response function},  $\hat{\theta}_{\epsilon,z_{\text{train}}}$, which describes what the optimal parameters \emph{would be} if we were to infinitesimally upweight  the loss (by $\epsilon$) on a specific point $z_\text{train}=(x_\ttrain, y_\ttrain)$ in the empirical risk.
This perturbed objective is defined as:
\vspace{-5pt}
\begin{equation}
    \hat{\theta}_{\epsilon,z_{\ttrain}} 
    = \arg\min_{\theta} \frac{1}{n}\sum_{i=1}^n \ell(h_\theta(x_i), y_i) 
    + \epsilon \ell(h_\theta(x_{\ttrain}), y_{\ttrain}),
\end{equation}
where the solution at $\epsilon = 0$ corresponds exactly to the original pre-trained parameters: $\hat{\theta}_{0, z_{\text{train}}} = \hat{\theta}$.
The IF 
measures the sensitivity of 
these pre-trained parameters 
by computing the first-order Taylor approximation (i.e., the derivative) of the response function with respect to $\epsilon$, at $\hat{\theta}$. 
Under standard regularity conditions, this can be computed using the Implicit Function Theorem~\citep{krantz2002implicit}.
Let $H_{\hat{\theta}} = \frac{1}{n}\sum_{i=1}^n \nabla^2_\theta \ell(h_{\hat{\theta}}(x_i), y_i)$ be the Hessian of the empirical risk evaluated at $\hat\theta$, then
\begin{equation}
    \mathrm{IF}_{\hat\theta}(z_{\ttrain})=
    \frac{d \hat{\theta}_{\epsilon,z_{\ttrain}}}{d\epsilon}\bigg|_{\epsilon=0}
    = - H_{\hat{\theta}}^{-1} \nabla_\theta \ell(h_{\hat{\theta}}(x_{\ttrain}), y_{\ttrain}).
\end{equation}
\textbf{Influential Training Samples on Test Prediction\;\;} 
Since $\mathrm{IF}_{\hat{\theta}}(z_{\text{train}})$ is a high-dimensional vector, it is often difficult to interpret directly. To obtain a more concrete measure, we convert this parameter-space influence into a scalar quantity by measuring its effect on a specific model output. This is done by projecting the influence vector onto the gradient of a chosen function, such as the loss or the logits for a test example $z_{\text{test}} = (x_\ttest, y_\ttest)$.
Applying the Chain Rule,
we can compute the scalar \emph{influence} of upweighting $z_{\text{train}}$ on the loss at $z_{\text{test}}$ as follows:
\begin{equation}
\label{eq:if_full}
    \mathcal{I}(z_{\ttrain}, z_{\text{test}})
    = - \nabla_\theta \ell(h_{\hat{\theta}}(x_{\text{test}}), y_{\text{test}})^\top
      H_{\hat{\theta}}^{-1}\\
    \quad \nabla_\theta \ell(h_{\hat{\theta}}(x_{\ttrain}), y_{\ttrain}).
\end{equation}
This provides an interpretable measure to trace predictions back to influential training samples.

\textbf{Influential Tokens on Test Prediction in Autoregressive Tasks\;\;}
In \emph{autoregressive} tasks, the loss function decomposes additively across tokens, which enables the direct computation of token-level influence. This additive structure permits the gradient and Hessian in the influence function to be similarly decomposed, allowing the influence of individual training tokens to be derived explicitly.
Let $\{x_1, \cdots,x_T\}$ to denote the $T$ tokens in $x_\ttrain$. Then, \eqref{eq:if_full} can be rewritten as
\begin{equation}
\label{eq:if_full_token}
\mathcal{I}(z_{\ttrain}, z_{\text{test}})
= - \nabla_\theta \ell(h_{\hat{\theta}}(x_{\text{test}}), y_{\text{test}})^\top
  H_{\hat{\theta}}^{-1}\\
\quad \nabla_\theta \sum_{t=1}^T\ell(h_{\hat{\theta}}(x_{t}), y_{t}).
\end{equation}
Thus, the per-token influence score is defined as \citep{grosse2023studying}:
\vspace{-2pt}
\begin{equation}
\label{eq:if_per_token}
\mathcal{I}_t(z_\text{train}, z_{\text{test}})
= -  \nabla_\theta \ell(h_{\hat{\theta}}(x_{\text{test}}), y_{\text{test}})^\top
  H_{\hat{\theta}}^{-1}\\
\quad \nabla_\theta \ell(h_{\hat{\theta}}(x_{t}), y_{t}).
\end{equation}
\begin{remark}[Problems with Existing Per-Token Influence]\label{remark:per-token-if}
However, the decomposition in \eqref{eq:if_per_token} is restricted to an autoregressive task and implicitly assumes that the tokens in each training sample are independent. This is violated in text, as tokens are highly correlated. Consequently, the influence score for a token captures not only its own effect but also the confounded effects of correlated tokens in its context. This entanglement breaks the core interpretation of the score as measuring the isolated effect of a single token, rendering the estimates unreliable. To address this, we propose augmenting the LLM with modified SAEs (Section \ref{sec:method}), which enable influence estimation in a structured latent space where these dependencies can be better controlled.
\end{remark}


\paragraph{Jacobian-Vector Products}
This is a key technical tool that we use.
Given a function $F:\mathbb{R}^n \to \mathbb{R}^m$ and a direction $v \in \mathbb{R}^n$,  
the JVP at $x\in\R^n$ is the \emph{directional derivative} of $F$ at $x$ along $v$: 
\begin{equation}\label{eq:def-jvp}
    \mathrm{JVP}(F, x, v) \; = \left.\frac{d}{d\varepsilon} F(x + \varepsilon v)\right|_{\varepsilon=0} 
    \;=\;  J_F(x)\,v,
\end{equation}
where $J_F(x)$ is the Jacobian of $F$ at $x$.  
Intuitively, it answers the question:  
\emph{“If I nudge the input by an infinitesimal step $\varepsilon v$, how does the output change to first order?”}

Modern automatic differentiation libraries (e.g., PyTorch, JAX, TensorFlow) can compute JVPs directly without materializing the full Jacobian.  
Instead, they propagate the perturbation $v$ forward through each primitive operation (forward-mode AD),  
making JVPs scalable to high-dimensional functions such as deep neural networks.

%% file: _s4_method.tex
\section{Methodology}\label{sec:method}
\begin{figure*}[t]
\centering
\includegraphics[width=\textwidth]{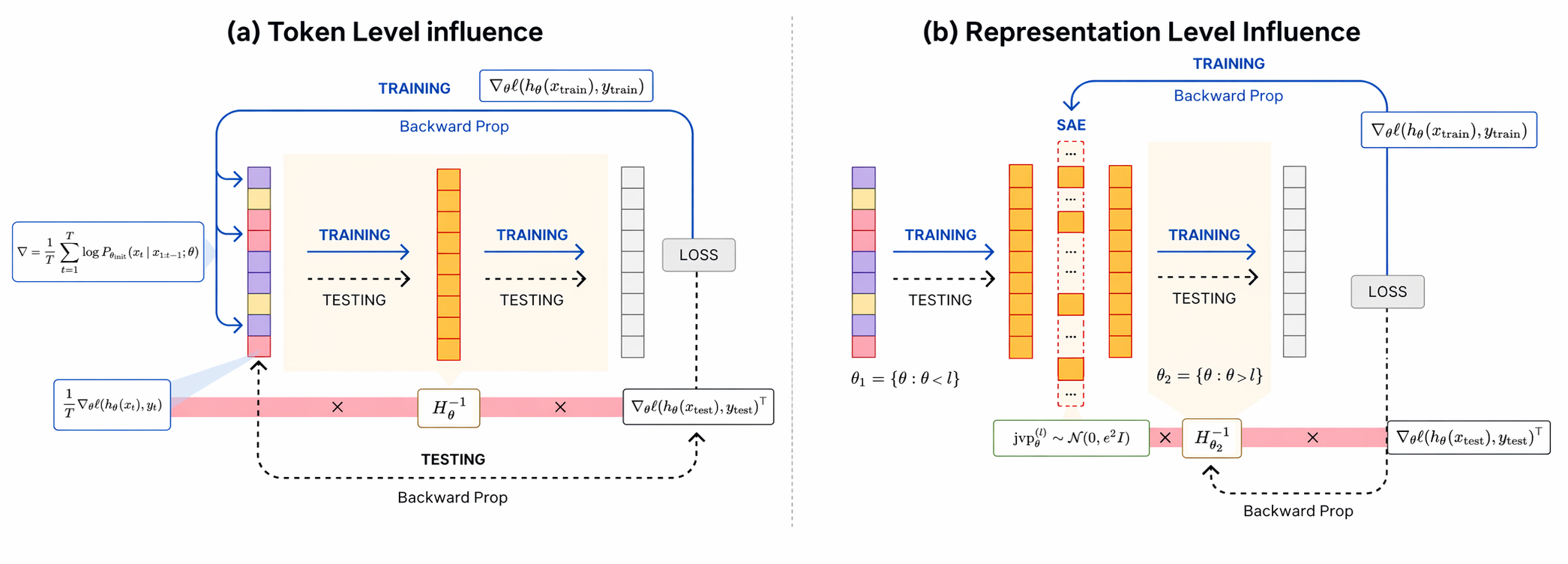}
\caption{
\textbf{Framework overview.} 
Traditional influence functions operate in the input space, assuming token independence and decomposable losses.  
Our method introduces a sparse autoencoder at an intermediate layer, splitting the model into upstream and downstream parts.  
Influence is then computed at the representation level using JVPs, enabling stable per-feature attributions and linking test predictions to interpretable sparse features.  
}
\label{fig:pipeline}
\end{figure*}

We now detail our framework that infers \emph{token-level} influence on test predictions via a latent mediation approach, 
enabling more reliable influence estimation for general prediction tasks. This section presents the core components of our approach: 1) augmenting LLMs with SAEs
to obtain more interpretable latent representations (Section~\ref{subsec:SAE}), 2) computing influence scores over these latent features rather than directly on input tokens (Section~\ref{subsec:latent_IF}), and 3) efficiently implementing this computation via Jacobian-vector products (Section~\ref{subsec:JVP}) while maintaining the ability to propagate attributions back to the input space. Figure~\ref{fig:pipeline} provides an overview of the complete framework.

\subsection{Augmenting LLM with Sparse Autoencoders
for Independent Features}\label{subsec:SAE}
We follow~\citet{bricken2023monosemanticity} and~\citet{gao2024scaling} to define a sparse autoencoder that maps input $x^l \in \mathbb{R}^d$ at layer $l$ into a sparse latent code $r \in \mathbb{R}^h$ through
\begin{align}
    r &=  \sigma(W_{\enc} (x^l-b_{\pre}) + b_{\enc}), \\
    \tilde{x}^l &= W_{\dec} r + b_{\pre},
    \vspace{-5pt}
\end{align}
where $W_{\enc} \in \R^{h\times d}$, $b_{\enc} \in \R^h$, $W_{\dec} \in \R^{d\times h}$, and $b_{\pre} \in 
\R^d$. 
The nonlinearity $\sigma(\cdot)$ is ReLU in classical settings
\citep{bricken2023monosemanticity} and TopK in modern settings \citep{gao2024scaling}.

Let $r_j$ be the activation of feature $j$, 
forming the basis for our feature-level influence analysis.

\begin{remark}[SAE for Improved Influence Estimation]\label{remark:SAE}
Classical influence functions assume independent samples, an assumption broken by token-level representations, where strong sequential correlations invalidate estimates. SAEs, by contrast, induce a latent representation comprised of approximately independent features,\footnote{To promote orthogonality, we also experimented with adding an orthogonality constraint. Please see Table~\ref{tab:orthogonality-ablation} for additional details. In our final visualization, we excluded the orthogonality penalty due to the accuracy sparsity tradeoff.}
each corresponding to a semantically meaningful concept. While these features are not strictly independent, their distributions are regularized toward comparable sparsity patterns---significantly closer to the independent structure assumptions underlying influence estimation. 
This alignment  makes SAE-based latents far more suitable for influence score interpretation than raw token-level attribution, where strong sequential dependencies violate core requirements of the influence framework. Although influence functions technically require features to be identically distributed for comparable scaling across components, this represents a second-order concern. We expect SAEs to produce latent representations with more comparable distribution scales than raw tokens, thereby providing more reliable influence estimates.
\end{remark}

\subsection{Influence Functions on Latent Representation}\label{subsec:latent_IF}

Classical influence functions applied directly to correlated text tokens (\eqref{eq:if_per_token}) are problematic due to strong sequential dependencies. To address this, we compute influence on \emph{latent features} rather than raw tokens. 
Consider an intermediate activation $h^{(l)}\in\mathbb{R}^{d_l}$ at layer $l$, which may correspond to the output of an attention head, MLP block, or residual stream in a transformer~\citep{elhage_mathematical_2021}. Mechanistic interpretability studies suggest that such representations often encode semantically meaningful features causally linked to final predictions~\citep{wang_interpretability_2023,meng_locating_2022}.
To attribute influence to individual neurons (latent coordinates) within the representation $r_\ttrain$ (corresponding to $h^{(l)}$), we must relate neuron-level effects directly to the \emph{training gradient} that appears in influence functions.

As illustarated in Figure~\ref{fig:pipeline}, we split the model parameters into two parts: $\theta = (\theta_1, \theta_2)$,
where $\theta_1=\{\theta:\theta_{<l}\}$ 
comprises all parameters \emph{up to and including layer $l$}, mapping a raw input sequence $x$ to an intermediate representation $r = h_{\theta_1}(x)$; 
$\theta_2=\{\theta:\theta_{>l}\}$
comprises the remaining parameters \emph{after layer $l$}, mapping $r$ to the final prediction.
By fixing $\theta_1$ and treating $r$ as the input, the IF can be computed with respect to $\theta_2$ alone—effectively attributing influence to the latent representation $r$ rather than the original tokens.

Let $r_\ttrain = h_{\theta_1}(x_\ttrain)$ and $r_\ttest = h_{\theta_1}(x_\ttest)$ denote the latent representation of the training and testing inputs, $x_\ttrain$ and $x_\ttest$, respectively.
Define the corresponding latent-space data points as $z_\ttrain^r = (r_\ttrain, y_\ttrain)$ and $z_\ttest^r = (r_\ttest, y_\ttest)$.

The \emph{representation-level influence function} is defined as:
\begin{equation}
\label{eq:rep_if}
\mathcal{I}(z^r_{\text{train}}, z^r_{\text{test}})
= - \left.\underbrace{\nabla_{\theta_2} \ell(h_{\theta_2}(r_{\text{test}}), y_{\text{test}})}_{=:\,g_\ttest} \right.^{\top}
  H_{\theta_2}^{-1}\\
\quad \underbrace{\nabla_{\theta_2} \ell(h_{\theta_2}(r_{\text{train}}), y_{\text{train}})}_{=:\,g_\ttrain},
\end{equation}
where $H_{\theta_2}$ is the Hessian of the training loss w.r.t. $\theta_2$.

A crucial \textbf{philosophical asymmetry} now arises: IFs measure how a \emph{training point} affects the loss on a \emph{test point}. The test point serves as a fixed evaluation context—we care about its loss, but we do not attribute influence to its internal structure. Consequently, while both $z^r_{\ttrain}$ and $z^r_{\ttest}$ are latent representations, our goal is to decompose the training-side gradient $g_\ttrain$ into contributions from individual latent coordinates of $r_\ttrain$.
In contrast, $g_\ttest$ requires no decomposition; it is obtained by a single backward pass through $\theta_2$ after computing $r_\ttest$ via $\theta_1$. 
To make this asymmetry explicit in notation, we use the notation
$\mathcal{I}^r(z^r_{\text{train}}, z_{\text{test}})$:
\begin{equation}\label{eq:rep_if_v2}
    \mathcal{I}^r(z^r_{\text{train}}, z_{\text{test}}) 
    = \mathcal{I}(z^r_{\text{train}}, 
    (h_{\theta_1}(x_\ttest), y_\ttest))
    = \mathcal{I}(z^r_{\text{train}}, z^r_{\text{test}}),
\end{equation}
where the superscript $r$ on $\mathcal{I}$ signals that attribution targets the \emph{training representation}.
The core challenge lies in attributing $g_\ttrain$ to individual latent coordinates of $r_\ttrain$ in a manner that reflects their actual computational attribution,\footnote{Decomposing $g_\ttest$ would answer a different question (e.g., “which test features make it sensitive to training data”). The Hessian $H_{\theta_2}$ depends solely on training data, and the perturbation is applied to the training point; thus, $r_\ttest$’s internal composition is irrelevant for the attribution we seek.} which we articulate in Section~\ref{subsec:integrated_gradients}.
Now, by mapping these influential features back to the specific words that \emph{activate} them, our explanations more faithfully capture the model’s internal reasoning—moving beyond isolated token attributions toward coherent, feature-driven interpretability.


\subsection{Feature-Level Attribution via Integrated Gradients}
\label{subsec:integrated_gradients}
Recall that our goal is to attribute the training-side gradient to individual latent coordinates of $r_\ttrain$. To make this dependence explicit and to evaluate it at different inputs, we define a function $G$ that any intermediate representation $r$ to the corresponding gradient on $\theta_2$:
\begin{equation}
G(r)\;=\;\nabla_{\theta_2}\ell\!\left(h_{\theta_2}(r),y_\ttrain\right),
\end{equation}
with the loss evaluated using the fixed training label $y_\ttrain$.
By construction, $g_\ttrain = G(r_\ttrain)$. 

In contrast to token attribution in autoregressive tasks, $G(r)$ is a single vector within the high-dimensional parameter space of $\theta_2$; consequently, it cannot be linearly separated into distinct components for individual neurons. 
A naive approach would be to consider the directional derivative $\partial G / \partial r^{(j)}$, which measures how an infinitesimal perturbation to neuron $j$ affects the gradient. While this captures local sensitivity, it does not tell us how much that neuron's \emph{activation} contributes to the actual value of $G(r)$ when transitioning from an inactive to an active state.
This issue is commonly addressed by defining contributions relative to a baseline representing the absence of the features
(e.g., integrated gradients~\citep{sundararajan_axiomatic_2017}, Shapley values~\citep{lundberg2017unified}). 
For sparse representations learned by SAEs, the natural baseline is
$r_0=\mathbf{0}$, corresponding to the manifold
where ``all features are inactive.'' We therefore consider the change $\Delta G = G(r_\ttrain) - G(\mathbf{0})$, which captures the effect of turning on the active features.

To decompose $\Delta G$ into per-neuron attributions, we adopt the axiomatic framework of integrated gradients~\citep{sundararajan_axiomatic_2017}.
This method attributes the output of a scalar model to its input features by integrating the gradients along a straight-line path from a baseline to the realized representation.
We parametrize this straight-line path by $\alpha \in [0,1]$, and, abusing the notation, denote it as $r(\alpha)$.
Because our baseline is the origin ($r_0=\mathbf{0}$), this straight-line path is the scaling function $r(\alpha) = \alpha r_\ttrain$.
By the chain rule of calculus, the rate of change of the gradient along this path is driven entirely by the representation itself:
\begin{equation}
\label{chain_rule_of_G}
\frac{d}{d\alpha}G(r(\alpha)) = J_G(\alpha r_\ttrain) \frac{dr(\alpha)}{d\alpha} = J_G(\alpha r_\ttrain) r_\ttrain.
\end{equation}

Applying the fundamental theorem of calculus, we integrate this derivative to express the total gradient change:
\begin{equation}\label{eq:grad_path_decomp}
G(r_\ttrain) - G(\mathbf{0}) = \int_0^1 J_G(\alpha r_\ttrain)\,r_\ttrain\,d\alpha,
\end{equation}
where $J_G(r)\in\mathbb{R}^{|\theta_2|\times d_l}$ is the Jacobian of $G$ with respect to $r$. Expanding the product coordinate-wise yields an exact additive latent decomposition:
\begin{equation}\label{eq:latent_grad_decomp_exact}
G(r_\ttrain)
=
G(\mathbf{0})
+
\sum_{j=1}^{d_l} r_\ttrain^{(j)}
\left(\int_0^1 J_G(\alpha r_\ttrain)\,e_j\,d\alpha\right).
\end{equation}
We provide a detailed proof of this decomposition in Appendix~\ref{app:path_integral_decomposition}.

Computing the exact integral is expensive.
To obtain a scalable estimator, we approximate the integral by evaluating the Jacobian at a particular point along the path, $r^\star$.
This represents a backward first-order Taylor expansion from the realized activations, defining our per-neuron latent contribution as:
\begin{equation}\label{eq:latent_contrib_def}
\Delta G_j
(r^\star
)
\;:=\;
r_\ttrain^{(j)}\,J_G(
r^\star
)\,e_j.
\end{equation}

By choosing $r^\star =r_\ttrain$, this yields a computationally-efficient approximation 
because it evaluates the Jacobian using activations already materialized during the standard forward pass.
Moreover, the multiplicative factor  $r_\ttrain^{(j)}$ guarantees that contributions correctly vanish for inactive features, satisfying a desirable \emph{activation} sensitivity property.

\begin{definition}[Neuron-Level Influence]\label{def:neuron-if}
Using \eqref{eq:latent_contrib_def} to adjust for activation attribution,
the influence score attributed to neuron $j$ in the training representation $r_\ttrain$ is defined as
\begin{equation}\label{eq:per_representation_if}
\mathcal{I}_j^r(z^r_{\ttrain}, z_{\ttest})
=
-\, g_{\ttest}^\top H_{\theta_2}^{-1}\,
\Delta G_j(r_\ttrain),
\end{equation}
where $g_{\ttest}=\nabla_{\theta_2}\ell(h_{\theta_1}(x_\ttest),y_\ttest)$ and $H_{\theta_2}$ is the downstream Hessian.
\end{definition}

This formulation attributes influence to a neuron only if (i) it is active in the representation ($r_\ttrain^{(j)} \neq 0$) and (ii) its activation induces a change in the downstream parameter gradient, as quantified by the scaled Jacobian column $J_G(r_\ttrain)\,e_j$. 
The sign convention ensures that positive influence corresponds to improved train--test gradient alignment. Consequently, the score provides a causal, influence-based measure of each neuron's contribution, offering a rigorous alternative to correlational metrics like raw activation magnitude~\citep{koh2017understanding,geiger_causal_2021}.

\input{_s4_3_repinf}

%% file: _s4_3_repinf.tex




\subsection{Scaling Influence From JVPs to Constant-Time Derivative Swapping}
\label{subsec:JVP}


In Definition~\ref{def:neuron-if}, the contribution of neuron $j$ relies on the Jacobian column $J_G(r_\ttrain)\,e_j$. For LLMs, materializing the full Jacobian $J_G \in \mathbb{R}^{|\theta_2| \times d_l}$ is memory-prohibitive. A standard circumvention is to compute this column implicitly using a Jacobian-vector product (JVP)~\citep{baydin2017automatic}. By the definition of the directional derivative, evaluating the Jacobian along the standard basis vector $e_j$ yields:
\begin{equation}\label{eq:jvp_grad}
J_G(r_\ttrain)\,e_j
=
\left.\frac{d}{d\varepsilon}G(r_\ttrain+\varepsilon e_j)\right|_{\varepsilon=0}
=
\mathrm{JVP}(G, r_\ttrain, e_j).
\end{equation}

Putting Eq.s (\ref{eq:latent_contrib_def})-(\ref{eq:jvp_grad}) together, we obtain
a computable form for the neuron-level influence:
\begin{equation}\label{eq:per_representation_if_jvp}
\mathcal{I}_j^r(z^r_{\ttrain}, z_{\ttest})
=
-\, g_{\ttest}^\top H_{\theta_2}^{-1}\,
r_\ttrain^{(j)}\,
\mathrm{JVP}(G, r_\ttrain, e_j).
\end{equation}
Here, the JVP term quantifies the downstream gradient sensitivity, while the activation factor ($r_\ttrain^{(j)}$) enforces that only active circuits contribute. This directly connects data-level influence attribution to feature-level monosemanticity~\citep{bricken2023monosemanticity}.
 
\noindent\textbf{The Computational Bottleneck.\;\;} 
While 
\eqref{eq:per_representation_if_jvp} avoids storing the full Jacobian,
its evaluation requires
computing a separate JVP for \emph{every} active feature $j$, making it computationally impractical.
Because $G$ maps to the high-dimensional parameter space $\theta_2$, each JVP requires an expensive forward-over-reverse differentiation pass ($\mathcal{O}(d_l)$ passes). For SAEs where
thousands of features may fire simultaneously across a sequence, iterating 
over neurons individually becomes
computationally intractable. 
We therefore need a method that computes all neuron influences in constant time, independent of the number of active features.

\noindent\textbf{Constant-Time Influence via Derivative Swapping.\;\;} 
The key insight is to exchange the order of differentiation and summation
(via Clairaut's Theorem),
enabling us to obtain all feature influences from a single backward pass.

Let $s_\ttest = H_{\theta_2}^{-1} g_\ttest$
and define
$P(r_\ttrain) \;=\; s_\ttest^\top 
g_\ttrain.
$
Since $s_\ttest$ is constant w.r.t.\ $r_\ttrain$, we have
\begin{equation}
\nabla_{r_\ttrain} P
=
\nabla_{r_\ttrain}
\Big(
s_\ttest^\top g_\ttrain
\Big),
\end{equation}
and the vector of latent influences is
\begin{equation}
\label{eq:final_vectorized_influence}
\vec{\mathcal{I}}^r(z^r_\ttrain, z_\ttest)
=
-
\Big(
\nabla_{r_\ttrain} P
\Big)
\odot
r_\ttrain,
\end{equation}
reducing complexity from $\mathcal{O}(d_l)$ to $\mathcal{O}(1)$ backward passes (compute $\nabla_{\theta_2}\ell$, then backpropagate $P$ to $r_\ttrain$).

This constant-time formulation enables simultaneous computation of influence for all latent features and batch samples, allowing the method to scale to billion-parameter models and sparse autoencoders with tens of thousands of features, which significantly improves the efficiency of the framework. The full proof is provided in Appendix~\ref{app:derivative_swap_proof}. For speedup and memory-saving analysis, we redirect readers to Appendix~\ref{app:scaling_1b}.

%% file: _s5_experiment.tex
\section{Experiments}
\label{sec:exp}



\begin{table}[t]
\centering
\caption{Finetuning results across base models. $\Delta$ denotes LoRA-SFT minus baseline accuracy.}
\label{tab:ft-results}
\vspace{2mm}
\small
\begin{tabular}{lcccccc}
\toprule
& \multicolumn{3}{c}{CommonsenseQA} & \multicolumn{3}{c}{OpenBookQA} \\
\cmidrule(lr){2-4} \cmidrule(lr){5-7}
Model & Pretrained & LoRA-SFT & $\Delta$ & Pretrained & LoRA-SFT & $\Delta$ \\
\midrule
Llama-3.2-1B & 28.99\% & 69.21\% & +40.22\% & 26.60\% & 70.80\% & +44.20\% \\
Llama-3.2-1B-IT & 50.61\% & 72.40\% & +21.79\% & 42.00\% & 72.80\% & +30.80\% \\
Qwen2.5-1.5B & 67.49\% & 74.45\% & +6.96\% & 64.80\% & 79.60\% & +14.80\% \\
Qwen2.5-1.5B-IT & 72.81\% & 74.28\% & +1.47\% & 69.40\% & 77.60\% & +8.20\% \\
\bottomrule
\end{tabular}
\vspace{-3mm}
\end{table}

We evaluate on three multiple-choice QA benchmarks: MedQA~\cite{han2023medalpaca}, OpenBookQA~\cite{mihaylov2018openbookqa}, and CommonsenseQA~\cite{talmor2019commonsenseqa}. All experiments start from task-finetuned models, and influence is measured with respect to the resulting task adapted model at inference time.
We use OpenBookQA and CommonsenseQA as the primary quantitative benchmark in the main text, and defer additional dataset results and full experimental details to the appendix (Appendix~\ref{app:exp-details}). We study small open-weight, instruction-tuned LLMs from the Llama and Qwen families ($\sim$1B parameters).

\textbf{Setup and Pipeline}\;\;
For a given task, the pipeline proceeds as follows:
(1) finetune a pretrained LLM on the task dataset (Sec~\ref{sec:ftandsae});
(2) insert an SAE at an intermediate layer and train it with frozen LLM weights to obtain sparse latent features (Sec~\ref{sec:ftandsae}); 
(3) pre-filter candidate training examples using gradient similarity to reduce influence computation cost and evaluate pre-filtering stability (Sec~\ref{sec:grad-prefilter}). 
(4) compute representation-level influence scores at the SAE layer (defined in Sec.~\ref{subsec:JVP})
to rank training examples and latent features, 
and presents the quantitative evaluation (Sec.~\ref{sec:quantitative}).

\begin{table*}[t]
\centering
\caption{SAE insertion results on OpenbookQA. $\Delta$ denotes \texttt{+SAE} minus finetuned accuracy.}
\label{tab:sae-512-64}
\begin{tabular}{lccccccc}
\toprule
& & \multicolumn{3}{c}{Selected layer} & \multicolumn{3}{c}{Best layer} \\
\cmidrule(lr){3-5} \cmidrule(lr){6-8}
Model & LoRA-SFT & Layer & \texttt{+SAE} & $\Delta$ & Layer & \texttt{+SAE} & $\Delta$ \\
\midrule
Llama-3.2-1B-IT & 72.80 & 11 & 70.40\% & -2.40\% & 13 & 72.60\% & -0.20\% \\
Qwen2.5-1.5B-IT & 77.60 & 17 & 76.80\% & -0.80\% & 17 & 76.80\% & -0.80\% \\
\bottomrule
\end{tabular}
\end{table*}

\begin{table*}[t]
\centering
\caption{Orthogonality statistics across representation spaces on OpenbookQA, Llama-3.2-1B. Lower is better for entanglement metrics, higher is better for rank and near-orthogonality.}
\label{tab:orthogonality-llama}
\vspace{-2mm}
\begin{tabular}{lccccccc}
\toprule
Space & Gram Abs $\downarrow$ & Gram Sq $\downarrow$ & Frob. Norm $\downarrow$ & Stable Rank $\uparrow$ & \% $< 0.1$ $\uparrow$ & Acc \\
\midrule
Text       & 0.121 & 0.039 & 0.198  & 5.35  & 64.9\%   & 72.6\\
Pre-Latent & 0.824 & 0.730 & 0.854  & 1.17  & 2.0\%    & 72.6\\
SAE Latent & \textbf{0.008} & \textbf{0.002} & \textbf{0.047}  & \textbf{25.02} & \textbf{98.67\%}  & \textbf{73.2}\\
\bottomrule
\end{tabular}
\vspace{-3mm}
\end{table*}

\begin{figure*}[t]
\centering
\begin{subfigure}[t]{0.49\linewidth}
  \centering
  \includegraphics[width=\linewidth]{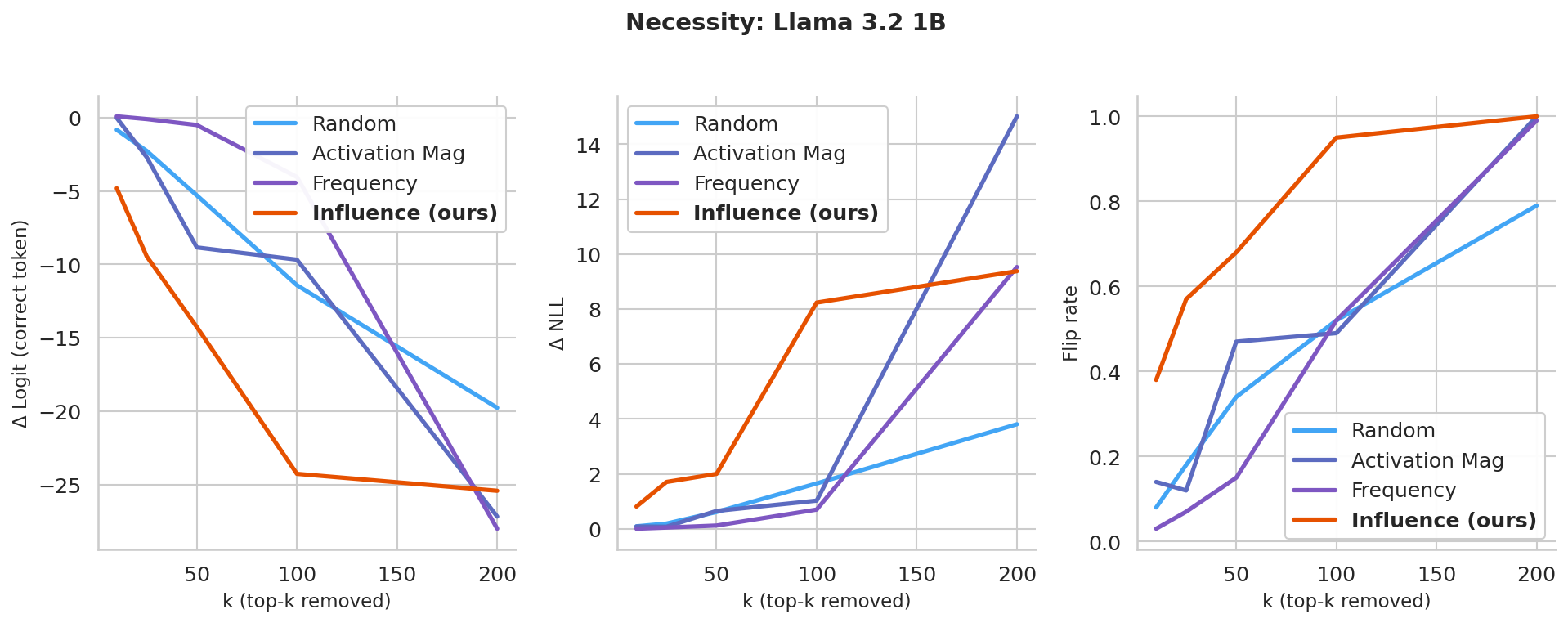}
  \caption{Llama-3.2-1B-Instruct (necessity).}
\end{subfigure}
\hfill
\begin{subfigure}[t]{0.49\linewidth}
  \centering
  \includegraphics[width=\linewidth]{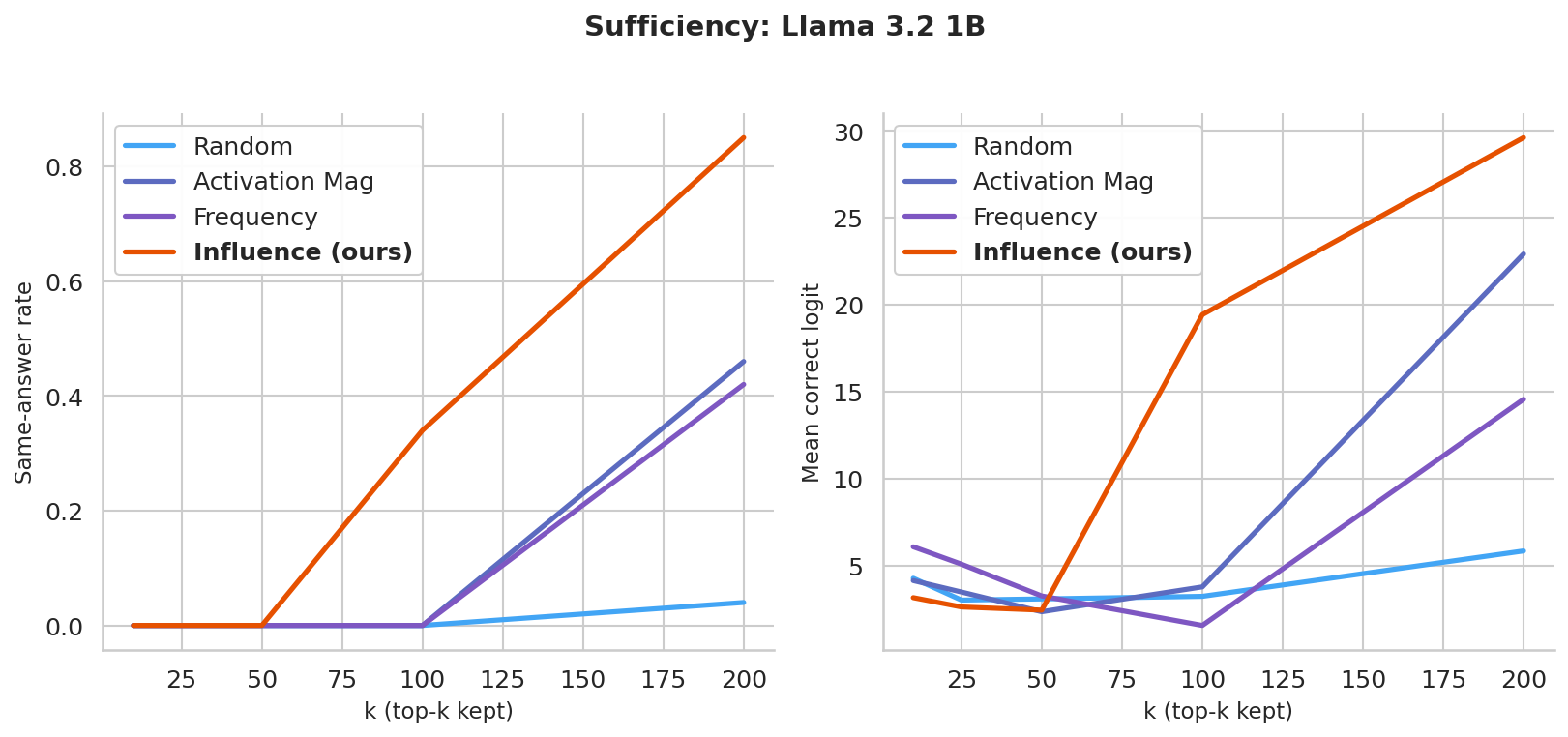}
  \caption{Llama-3.2-1B-Instruct (sufficiency).}
\end{subfigure}

\vspace{2mm}

\begin{subfigure}[t]{0.49\linewidth}
  \centering
  \includegraphics[width=\linewidth]{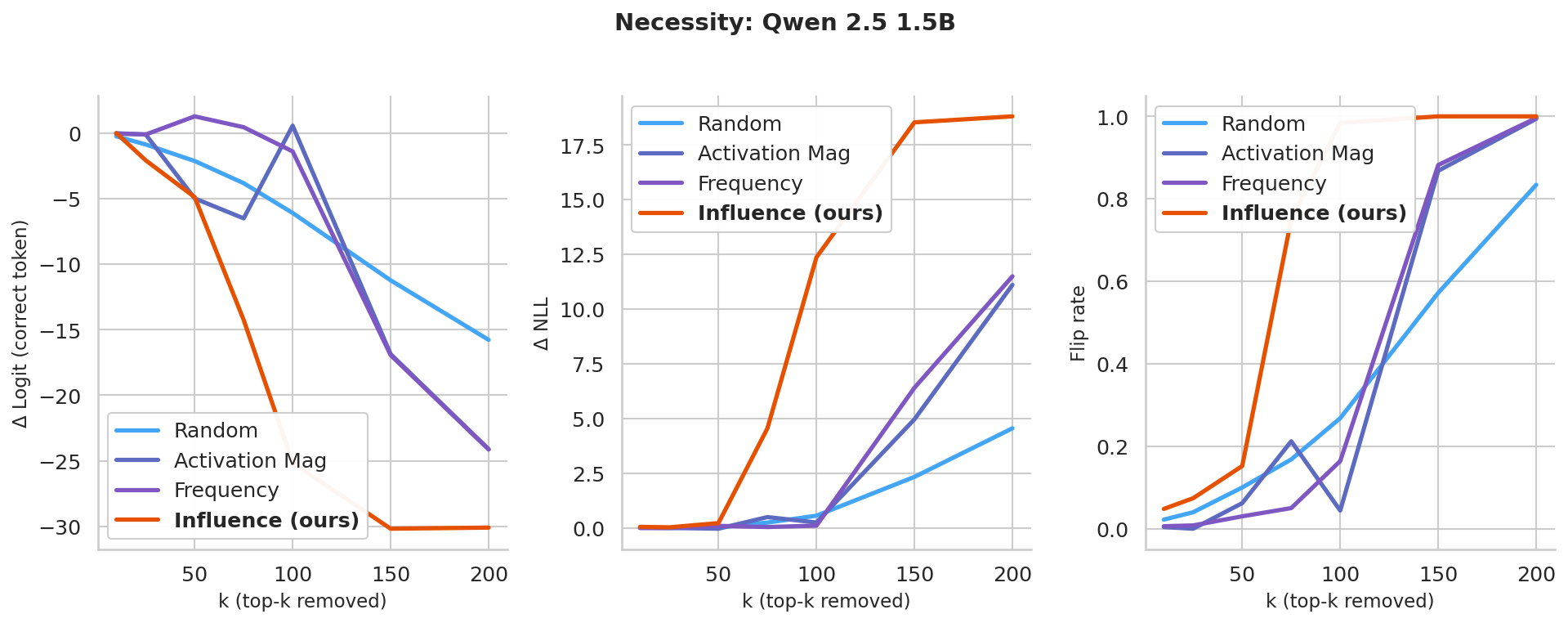}
  \caption{Qwen2.5-1.5B-Instruct (necessity).}
\end{subfigure}
\hfill
\begin{subfigure}[t]{0.49\linewidth}
  \centering
  \includegraphics[width=\linewidth]{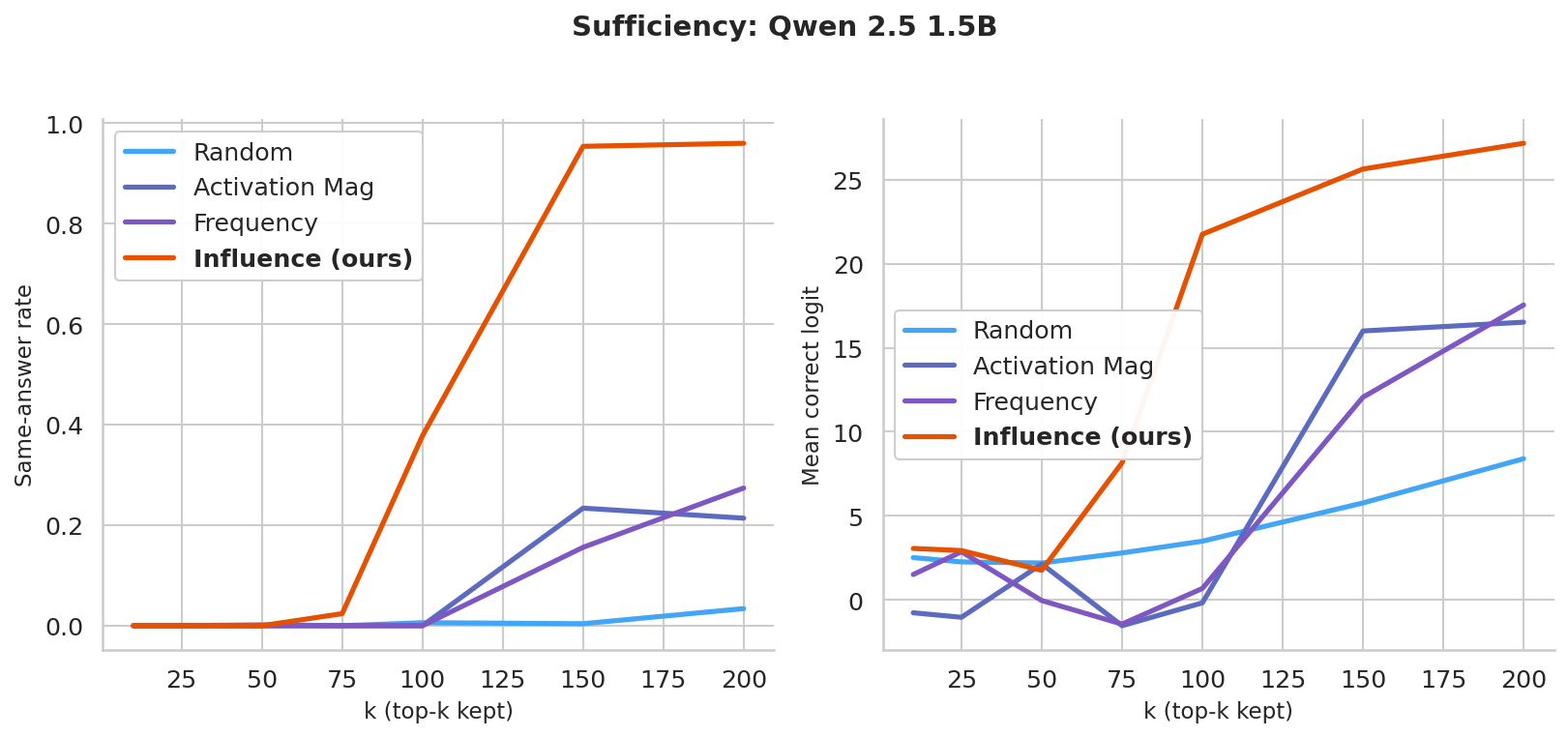}
  \caption{Qwen2.5-1.5B-Instruct (sufficiency).}
\end{subfigure}
\caption{Necessity and sufficiency tests on OpenbookQA. For necessity, we rank and remove top-$k$ SAE features using RepInf and three baselines, then report $\Delta$logit, $\Delta$NLL, and flip rate after masking. For sufficiency, we keep only top-$k$ features and report the retained correct logit and correct rate.}
\label{fig:necessity_sufficiency_openbook}
\end{figure*}

\vspace{-5pt}
\subsection{Task Performance After Finetuning and SAE Insertion}
\vspace{-5pt}
\label{sec:ftandsae}
A basic requirement for practical auditing is that 1) finetuning yields meaningful task-specific updates and 2) inserting the SAE layer does not significantly degrade the task performance.
We therefore report (i) pretrained and LoRA-SFT accuracy across base models in Table~\ref{tab:ft-results} and (ii) the accuracy change from inserting an SAE at the chosen layer in Table~\ref{tab:sae-512-64}. We additionally report full sweeps over SAE insertion layers across datasets and models in Appendix~\ref{app:sae-results}, showing that performance is sensitive to the placement of the representation bottleneck. 

\textbf{LoRA Supervised Finetuning.}\;\;
Table~\ref{tab:ft-results} shows that LoRA-SFT yields substantial gains across all base models, with the largest improvements for Llama-3.2-1B (+44.2\%/40.22\%) and Llama-3.2-1B-IT (+30.8\%/21.79\%). Qwen2.5-1.5B starts from a stronger baseline and thus shows smaller but still meaningful gains. These results confirm that finetuning produces large, measurable task specific updates, which is a prerequisite for downstream influence analysis.

\textbf{SAE insertion.}\;\;
Table~\ref{tab:sae-512-64} shows that inserting an SAE at an intermediate layer largely preserves finetuned performance: Qwen2.5-1.5B-IT drops by only 0.8 points at its selected (and best) layer, whereas Llama-3.2-1B-IT drops by 2.4 points at the selected layer. However, the ``best layer'' results indicate that most of Llama’s degradation can be recovered by a nearby layer choice (down to a 0.2 point drop), suggesting that the main cost comes from bottleneck placement rather than the SAE mechanism itself; layer selection is therefore a key hyperparameter for balancing interpretability and accuracy.

\textbf{Layer selection logic.}\;\;
Rather than selecting the globally best post-insertion layer, we choose the best layer within the middle half of the network (late layers can be less informative): layers 4--12 for Llama-3.2-1B-Instruct and 7--22 for Qwen2.5-1.5B-Instruct. We report both the \emph{selected} and \emph{best} layers, and provide the full sweep in the Appendix~\ref{app:sae-results}.


\subsection{Latent Space Orthogonality Analysis} 
We evaluate representation disentanglement by comparing feature orthogonality across three spaces: input text embeddings (\textbf{text}), dense pre-latent activations (immediately before SAE insertion, \textbf{Pre-Latent}), and the $k$-sparse SAE latent space (\textbf{SAE Latent}). Following prior work, we summarize orthogonality using the off-diagonal Gram statistics (mean absolute value, mean squared value, and Frobenius norm), the fraction of near orthogonal feature pairs with correlation $|\rho|<0.1$, and stable rank $||A||_F^2 / ||A||_2^2$~\citep{fel2025archetypal}. Table~\ref{tab:orthogonality-llama} shows a clear separation: pre-SAE latents are highly entangled with stable rank 1.17, 2.0\% near-orthogonal pairs, which says that most of the features are highly entangled with each other. Whereas SAE latents with high latent dimensions are substantially more disentangled with stable rank 25.02 and 98.67\% near orthogonal pairs everywhere, which proves that latents are naturally highly entangled even without proper guidance or constraints.
Then, compared with the main objective text embeddings that we claim to try to disentangle in this paper, the text embeddings on average have a stable rank of 5.35, 64.9\% near-orthogonal pairs, the SAE recovers more orthogonal directions while reducing pairwise correlation magnitude, supporting the use of sparse latents for influence estimation.
The specification of the SAE latent size and an ablation of the latent space orthogonality are included in Appendix~\ref{app:orth-ablation}.

\vspace{-5pt}
\subsection{Gradient Pre-Filtering for Scalable Influence Computation}
\label{sec:grad-prefilter}
\vspace{-5pt}
Exact influence computation over the full training set is expensive.
We therefore pre-filter training examples using gradient similarity and retain only the top 1\%--10\% candidates per test example.
Concretely, for a test example $z_{\mathrm{test}}$ and each training example $z_i$, we score
\begin{equation}
    s_i \;=\; \left\langle \nabla_{\theta}\mathcal{L}\!\left(z_i;\theta\right),\, \nabla_{\theta}\mathcal{L}\!\left(z_{\mathrm{test}};\theta\right) \right\rangle,
\end{equation}
and keep $\mathcal{C}_K(z_{\mathrm{test}})=\mathrm{TopK}_{i}(s_i)$.
We defer additional experiments on filtering thresholds and stability to Appendix~\ref{app:filtering}.


\begin{figure*}[t]
\centering
\begin{subfigure}[t]{0.49\linewidth}
  \centering
  \includegraphics[width=\linewidth]{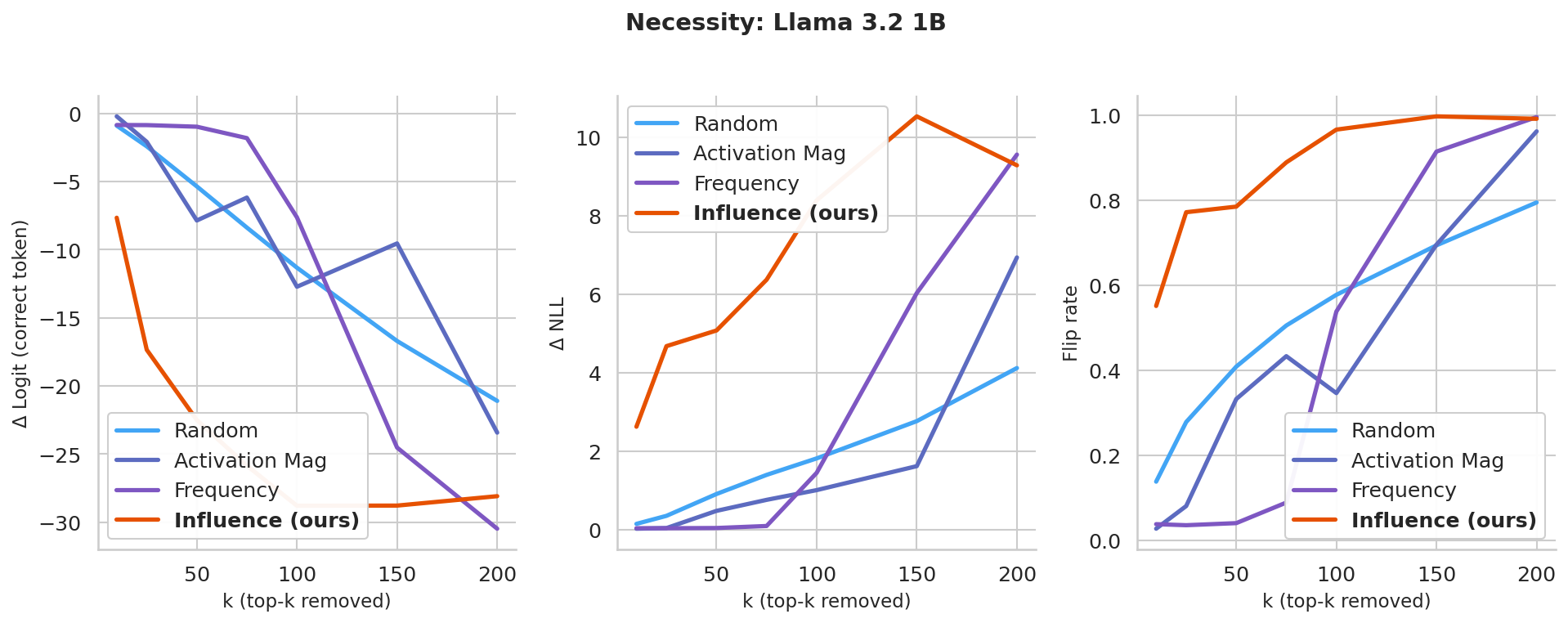}
  \caption{Llama-3.2-1B-Instruct (necessity).}
\end{subfigure}
\hfill
\begin{subfigure}[t]{0.49\linewidth}
  \centering
  \includegraphics[width=\linewidth]{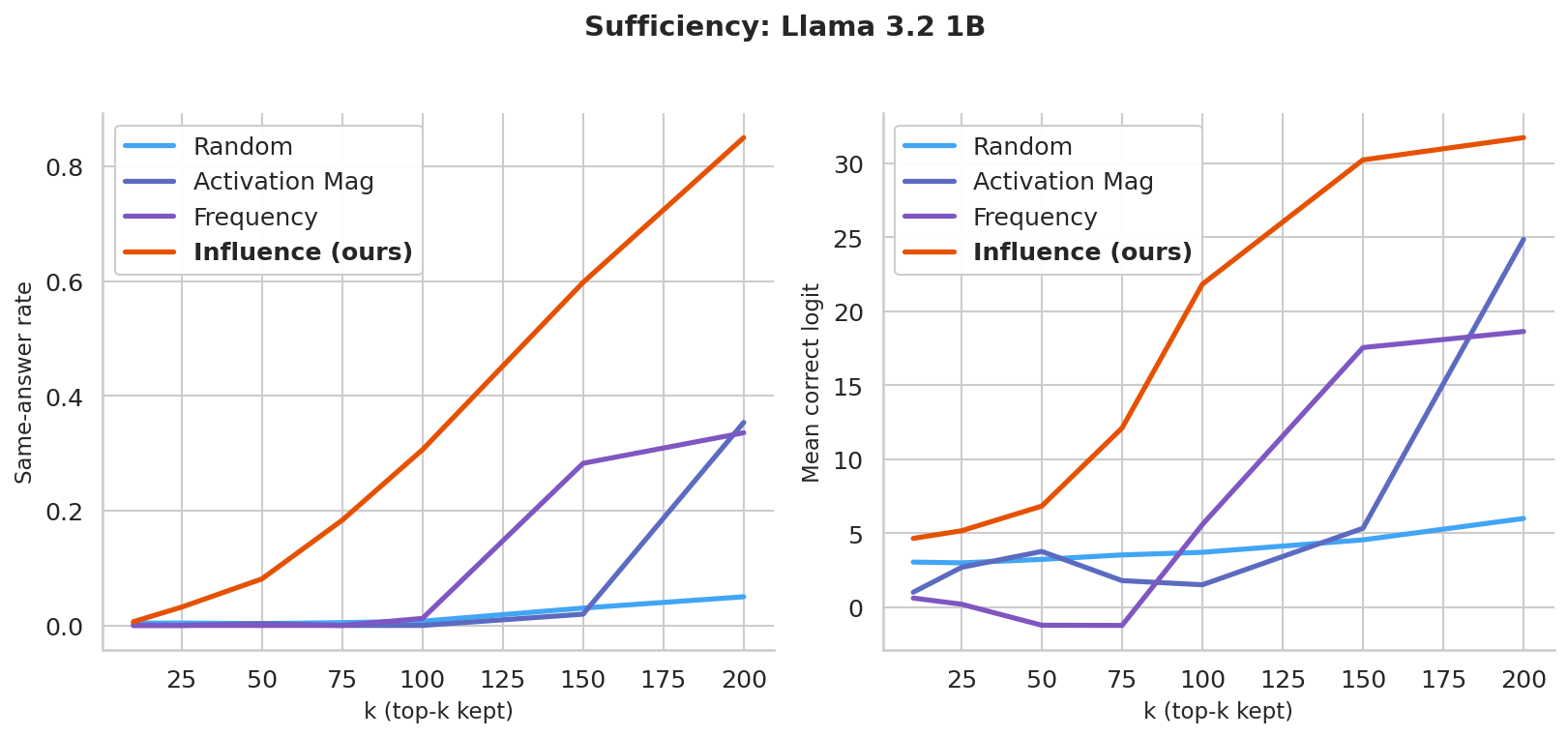}
  \caption{Llama-3.2-1B-Instruct (sufficiency).}
\end{subfigure}

\vspace{2mm}

\begin{subfigure}[t]{0.49\linewidth}
  \centering
  \includegraphics[width=\linewidth]{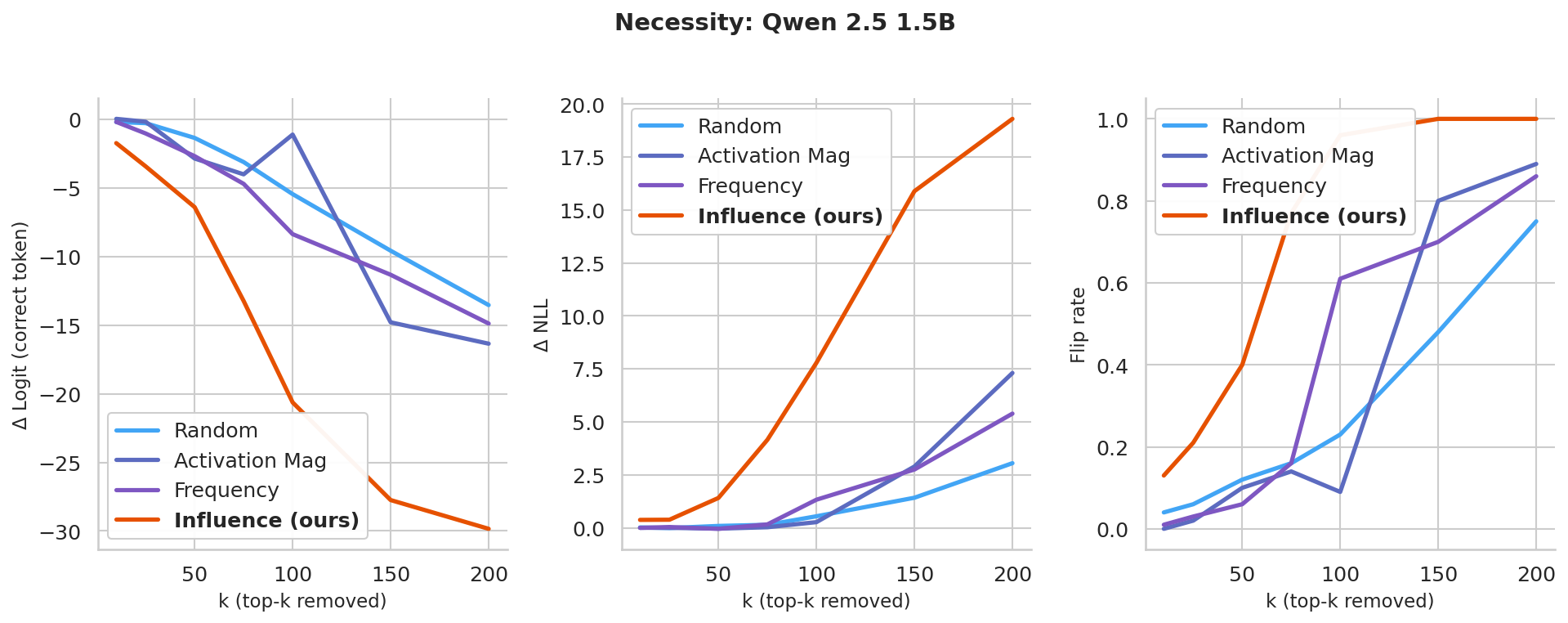}
  \caption{Qwen2.5-1.5B-Instruct (necessity).}
\end{subfigure}
\hfill
\begin{subfigure}[t]{0.49\linewidth}
  \centering
  \includegraphics[width=\linewidth]{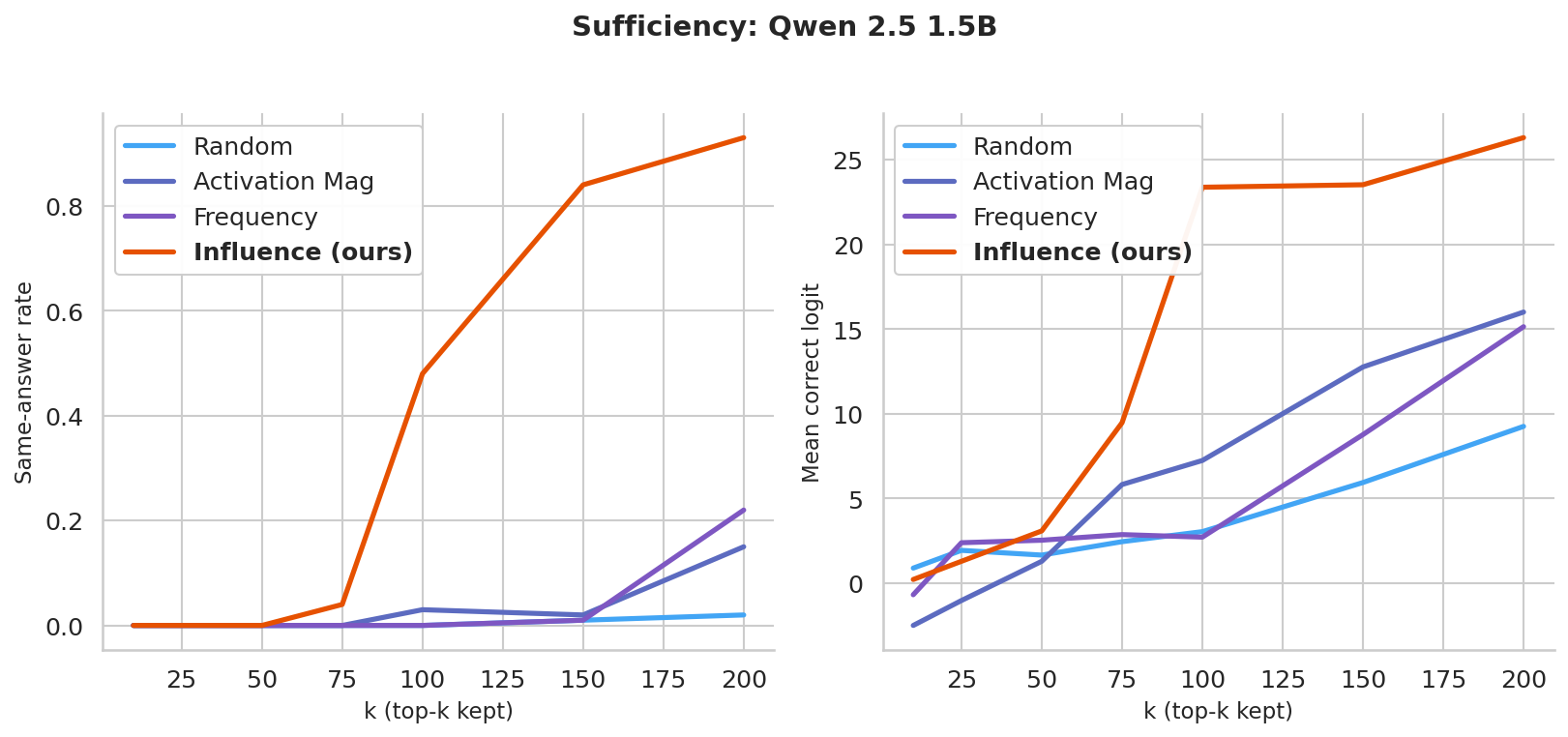}
  \caption{Qwen2.5-1.5B-Instruct (sufficiency).}
\end{subfigure}
\caption{CommonsenseQA necessity (remove top-$k$ features) and sufficiency (keep top-$k$ features) tests comparing influence-selected features to baselines.}
\label{fig:necessity_sufficiency_commonsenseqa}
\end{figure*}

\begin{figure}[t]
\centering
\begin{subfigure}[t]{\linewidth}
  \centering
  \includegraphics[width=0.8\linewidth]{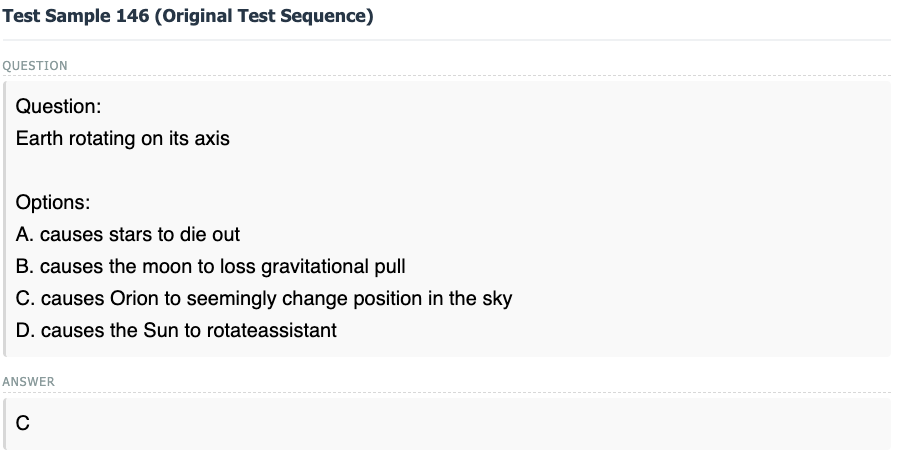}
  \caption{Original test input (token-level influence).}
\end{subfigure}

\begin{subfigure}[t]{\linewidth}
  \centering
  \includegraphics[width=0.8\linewidth]{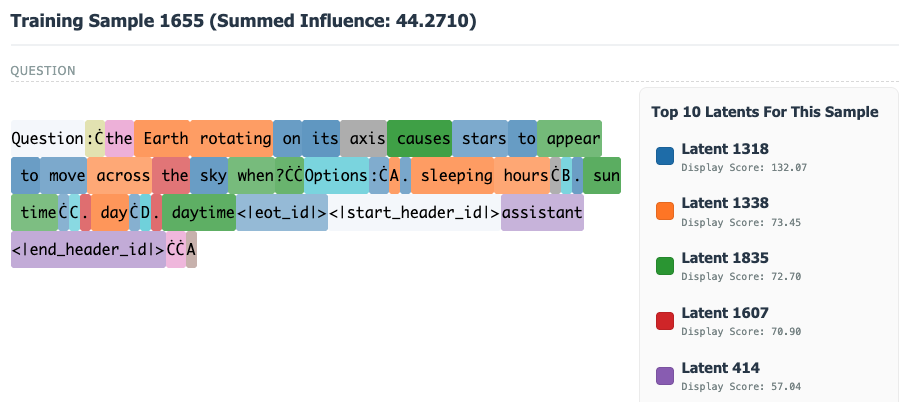}
  \caption{Most influential training sample (token-level influence).}
\end{subfigure}
\begin{subfigure}[t]{0.8\linewidth}
  \centering
  \includegraphics[width=0.9\linewidth]{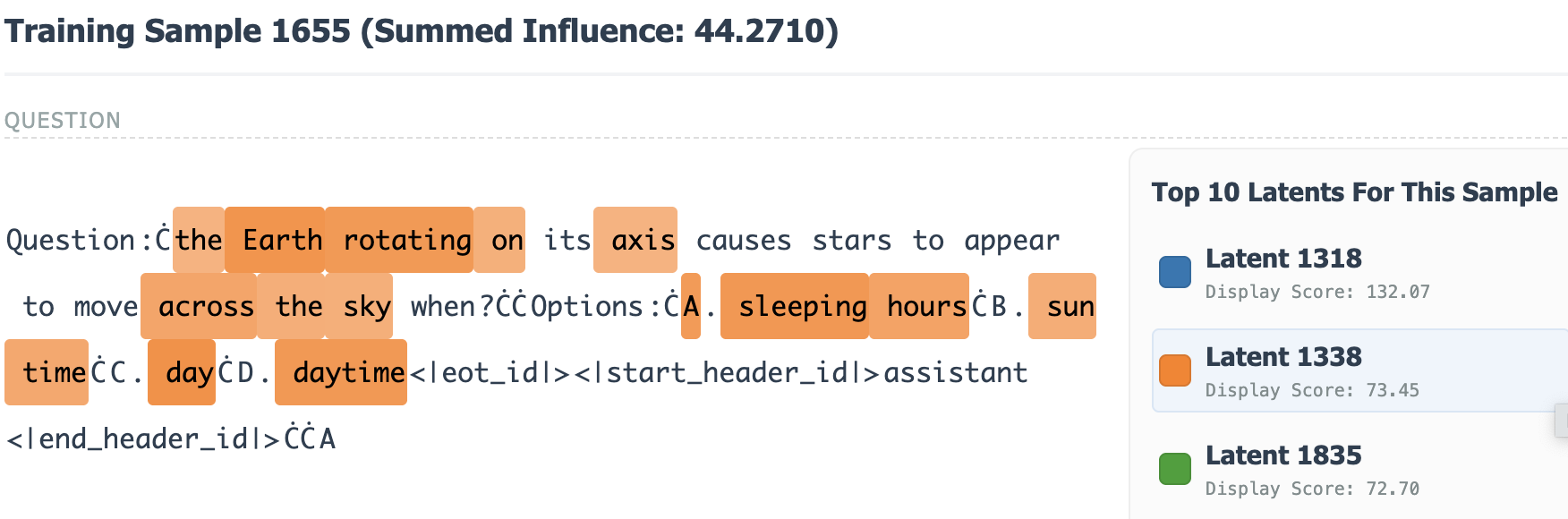}
  \caption{Individual feature 1338.}
  \label{fig:case1_latent}
\end{subfigure}
\caption{Token-level influence visualizations for a representative OpenBookQA test question (top), its most influential training example (middle), and its top influential token heatmaps (bottom). Colors denote the dominant influential latent feature and intensities denote per-token influence.}

\label{fig:heatmap-case1}
\end{figure}
\subsection{Quantitative Evaluation: Necessity and Sufficiency Tests}
\label{sec:quantitative}
To evaluate the faithfulness of sparse feature influence, we run two deletion/insertion style tests. For each evaluation sequence, we record the baseline prediction and correct-token statistics, then apply a binary mask at the SAE layer using a chosen ranking over latent features.

Our core signal is the representation level influence matrix $\mathrm{IFR}\in\mathbb{R}^{N\times H}$ for a test sample, where $N$ is the number of retained training examples after gradient pre filtering and $H$ is the SAE latent dimension; $\mathrm{IFR}[i,k]$ measures the influence of training example $i$ on latent feature $k$. We aggregate over training examples (mean over $i$) to obtain a test conditioned importance vector $s\in\mathbb{R}^{H}$ with $s_k=\frac{1}{N}\sum_{i=1}^{N}\mathrm{IFR}[i,k]$, and take the top-$k$ features under $s$ as the \emph{most influential} for that test sample.

We compare our influence score against three simple baselines: (i) ranking by activation magnitude, (ii) ranking by feature frequency, and (iii) a random set. Concretely, for each method we induce an ordering over latent dimensions and construct:

\begin{itemize}
    \item \textbf{Necessity (Remove Top-\textit{k}):} A mask that zeroes out the top-\textit{k} features, leaving other features intact.
    \item \textbf{Sufficiency (Keep Top-\textit{k}):} A mask that preserves only the top-\textit{k} ranked features, zeroing all others.
\end{itemize}

We evaluate both settings across multiple values of \textit{k} = \{25,50,100,125,150,175,200\}. For necessity, we report the average change in the correct token’s logit ($\Delta$logit), the average change in negative log likelihood ($\Delta$NLL), and prediction flip rate relative to the baseline. For sufficiency, we report the same answer rate and the retained mean correct logit under masking.


Figures~\ref{fig:necessity_sufficiency_openbook} and ~\ref{fig:necessity_sufficiency_commonsenseqa} demonstrate a consistent and quantitative separation between influence-based ranking and all baselines across both Llama-3.2-1B and Qwen2.5-1.5B.

\textbf{Necessity.}\;\;
Under the remove top-$k$ intervention, influence produces the steepest and most systematic degradation. On both models and both datasets, removing as few as 50 - 100 influence-ranked features induces a large negative shift in the correct token logit, a substantial increase in NLL, and a rapid rise in flip rate, often approaching saturation at moderate $k$. In contrast, random masking leads to gradual degradation, while activation magnitude and frequency exhibit intermediate behavior but consistently weaker impact. The approximately monotonic dependence of $\Delta$logit, $\Delta$NLL, and flip rate on $k$ suggests that influence induces a meaningful global ordering over latent features in terms of their causal contribution.

\textbf{Sufficiency.}\;\;
The keep top-$k$ experiment exhibits the complementary pattern. Influence retains substantially higher same answer rates and larger correct logits at every $k$. Notably, a relatively small subset of influence-ranked features suffices to recover a large fraction of the original predictive confidence, whereas random and heuristic rankings require many more features and still fail to match the retained performance. This indicates that predictive information is concentrated in a compact set of influence identified latent directions.

\textbf{Interpretation.}\;\;
Taken together, the dual behavior: sharp degradation under removal and strong recovery under retention, provides evidence that influence ranking captures features that are not merely highly active or frequent, but structurally implicated in the model’s decision. The consistency of this pattern across two distinct architectures further suggests that the effect reflects a representation level property rather than model specific artifacts.

\subsection{Qualitative Attribution: Linking Predictions to Training Evidence}
\label{sec:case-studies}
\vspace{-5pt}
We now present case studies showing how our method links a test-time prediction to specific training evidence.
For each case, we report the model prediction, the most influential training example, and token-level heatmaps obtained by projecting representation-level influence back to the input space.
\vspace{-2mm}
\paragraph{Case study protocol.}
We focus on test instances that are answered incorrectly by the pretrained model but correctly after finetuning, so that the resulting attributions reflect task-specific learning rather than generic priors. We run the same procedure for both Llama-3.2-1B-Instruct and Qwen2.5-1.5B-Instruct; for readability, the main text shows one representative example, while additional cases, including Qwen-based results, are deferred to Appendix~\ref{app:case-studies}. We additionally provide a heatmap visualization under reasoning augmented training in Case Study D (Figure~\ref{fig:heatmap-case4}), illustrating potential model behavior change when additional reasoning is provided. The model setup used in this subsection is included in Appendix~\ref{app:exp-details}.

\paragraph{Case Study A: Representation-level influence highlights shared astronomical reasoning cues.}
Figure~\ref{fig:heatmap-case1} shows an OpenBookQA question asking why Earth’s rotation on its axis makes Orion appear to change position in the sky. RepInfLLM retrieves a highly influential training example with summed influence 44.27: ``the Earth rotating on its axis causes stars to appear to move across the sky when?'', whose wording differs from the test question but encodes the same causal relation. The dominant latent features concentrate on the phrase group ``Earth rotating,'' ``on its axis,'' ``causes stars,'' ``appear,'' ``move,'' and ``sky,'' rather than on isolated lexical overlaps. This pattern suggests that the prediction is supported by a shared representation of the concept ``Earth's rotation causes apparent celestial motion,'' rather than by memorizing an answer string or matching option labels. More broadly, the example illustrates the value of representation-level influence: it surfaces semantically aligned training evidence and highlights the concept-bearing tokens that mediate the model's prediction.

We further inspect individual high influence latent features for the retrieved training sample. Latent 1338 primarily activates the concept bearing phrase “Earth rotating on its axis” together with “across the sky,” matching the causal astronomy relation required by the test question. Latent 1835 captures a complementary pattern involving “Earth,” “causes,” “appear,” and temporal answer choices such as “day” and “daytime.” Together, these latents suggest that the retrieved training example supports the prediction through multiple semantically coherent components: the physical mechanism of Earth’s rotation and the question-specific temporal framing. This provides more fine-grained evidence than a single aggregate influence score.

%% file: _s6_conclusion.tex
\section{Discussion}
\label{sec:discussion}

In this work, we introduce a novel interpretability framework based on IF that is applicable to any prediction task. Our proposed methods scale to large models with reasonable runtimes. Methodologically, our key contribution is the integration of SAEs into the LLM during fine-tuning, enabling the computation of influence scores over approximately orthogonal latent representations. By projecting these latent attributions back to the input space, our framework yields human-interpretable insights while preserving the theoretical soundness of influence estimation. Ablation studies demonstrate that our method achieves high necessity and sufficiency, confirming its ability to isolate the most influential latents driving a model's prediction on a given test sample. Beyond the current scope, this approach holds promise for multimodal settings, where it could serve as a diagnostic tool to interpret model behavior, identify failure modes, and assess performance bottlenecks across heterogeneous data modalities—offering a principled pathway toward more transparent and trustworthy multimodal systems.

However, several limitations remain. First, our method inherits the local nature of influence functions. The influence score approximates the effect of infinitesimal perturbations around a fixed trained model, and therefore should not be interpreted as exact leave-one-out retraining. It may miss nonlinear training effects, feature formation, or global representation changes that would occur under finite data removal or additional finetuning. Our method improves where the influence is measured, but it does not eliminate this fundamental approximation.

A second limitation is that SAE latents are only approximately disentangled. Our orthogonality analysis shows that SAE representations are much less entangled than dense pre-latent activations, but orthogonality does not guarantee semantic independence or causal modularity. Some latent features may still mix multiple concepts, and some concepts may be distributed across several features. Thus, the influence scores should be interpreted as attribution over a more structured latent basis, not as proof of fully independent causal concepts. Furthermore, there is a trade-off between the latent dimension size (reflected by the orthogonality score) and the interpretability of the final heatmap, which we plan to investigate in future work. 

Third, token-level visualizations are derived projections from latent influence, not the primary causal quantity. The strongest attribution produced by RepInfLLM is at the level of training examples and SAE features. Mapping these features back to input tokens makes the explanation more readable, but this projection depends on activation patterns and can blur cases where one feature is activated by multiple correlated tokens. Furthermore, the context in our datasets is relatively short, which may make it difficult for humans to achieve full interpretability of the latents. Future work using longer contexts could help address this limitation.

Finally, we do not report direct comparisons with standard data attribution methods because most prior
approaches operate in input space and evaluate instance-level rankings. In contrast, our framework
estimates representation-level influence in a sparse latent space and only subsequently projects
attributions back to text; conventional ground-truth protocols therefore do not transfer directly.

%% file: _s7_acknowledgement.tex
\section*{Acknowledgments}
We thank AWS for providing compute resources that supported prototyping of the framework. We also thank the NVIDIA Academic Grant Program Award for providing 32K A100 GPU-hours on Brev, which enabled us to scale up the experimental results.

%% file: _sa_appendix.tex
\section{Full Experimental Results}
This appendix contains the full results that the main experimental results haven't shown fully.

\subsection{SAE Layer Sweep and Training Results}
\label{app:sae-results}

We sweep SAE insertion layers and SAE configurations across benchmarks. Accuracy is typically best preserved in intermediate layers, and can degrade when inserting too early (low-level representations) or too late (near task-head computations).
In all plots, the \emph{shaded region} denotes the \emph{middle-half} of the network layers considered in the main experiments (Section~\ref{sec:ftandsae}). Concretely, for a model with $L$ transformer blocks, the middle-half range is
\[
\mathcal{L}_{\mathrm{mid}} \;=\; \{\lceil L/4\rceil, \ldots, \lfloor 3L/4\rfloor\}.
\]
The \emph{red dot} marks the best recovered accuracy (smallest drop) over the full sweep. The \emph{selected layer} is the best layer \emph{restricted} to $\mathcal{L}_{\mathrm{mid}}$.

\begin{figure*}[t]
\centering
\begin{subfigure}[t]{0.32\textwidth}
  \centering
  \includegraphics[width=\linewidth]{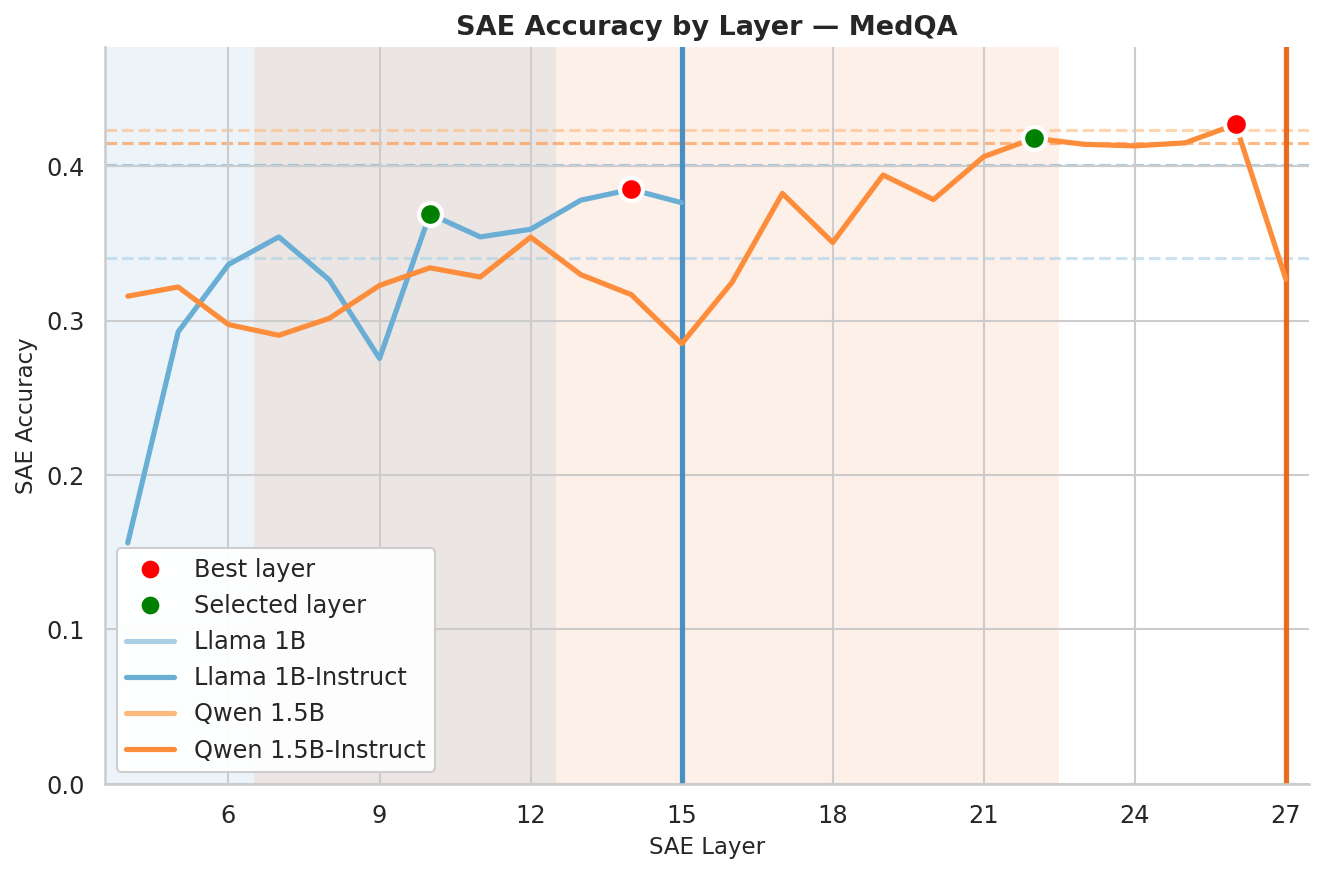}
  \caption{MedQA.}
  \label{fig:sae-layer-sensitivity-medqa}
\end{subfigure}
\hfill
\begin{subfigure}[t]{0.32\textwidth}
  \centering
  \includegraphics[width=\linewidth]{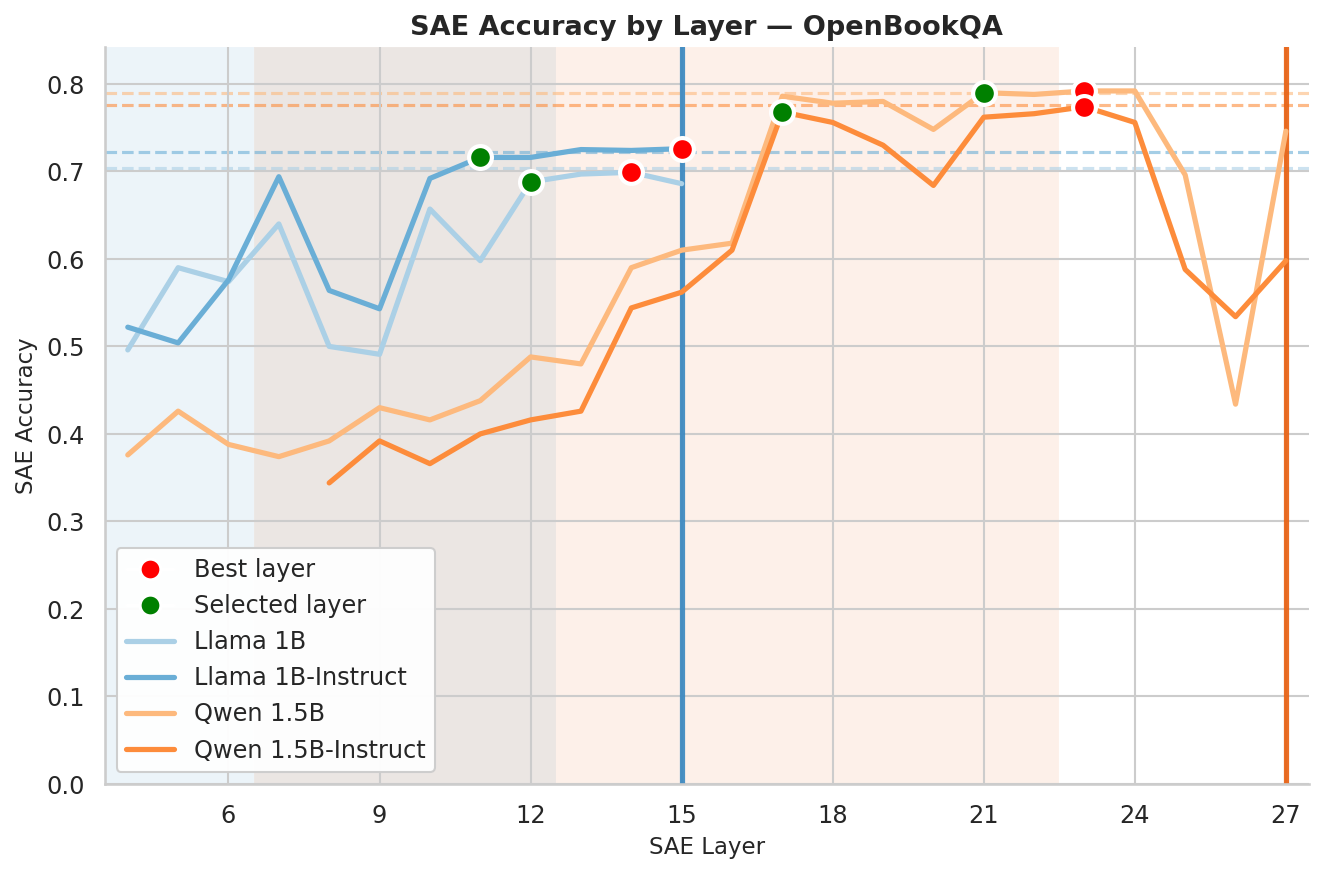}
  \caption{OpenBookQA.}
  \label{fig:sae-layer-sensitivity-openbookqa}
\end{subfigure}
\hfill
\begin{subfigure}[t]{0.32\textwidth}
  \centering
  \includegraphics[width=\linewidth]{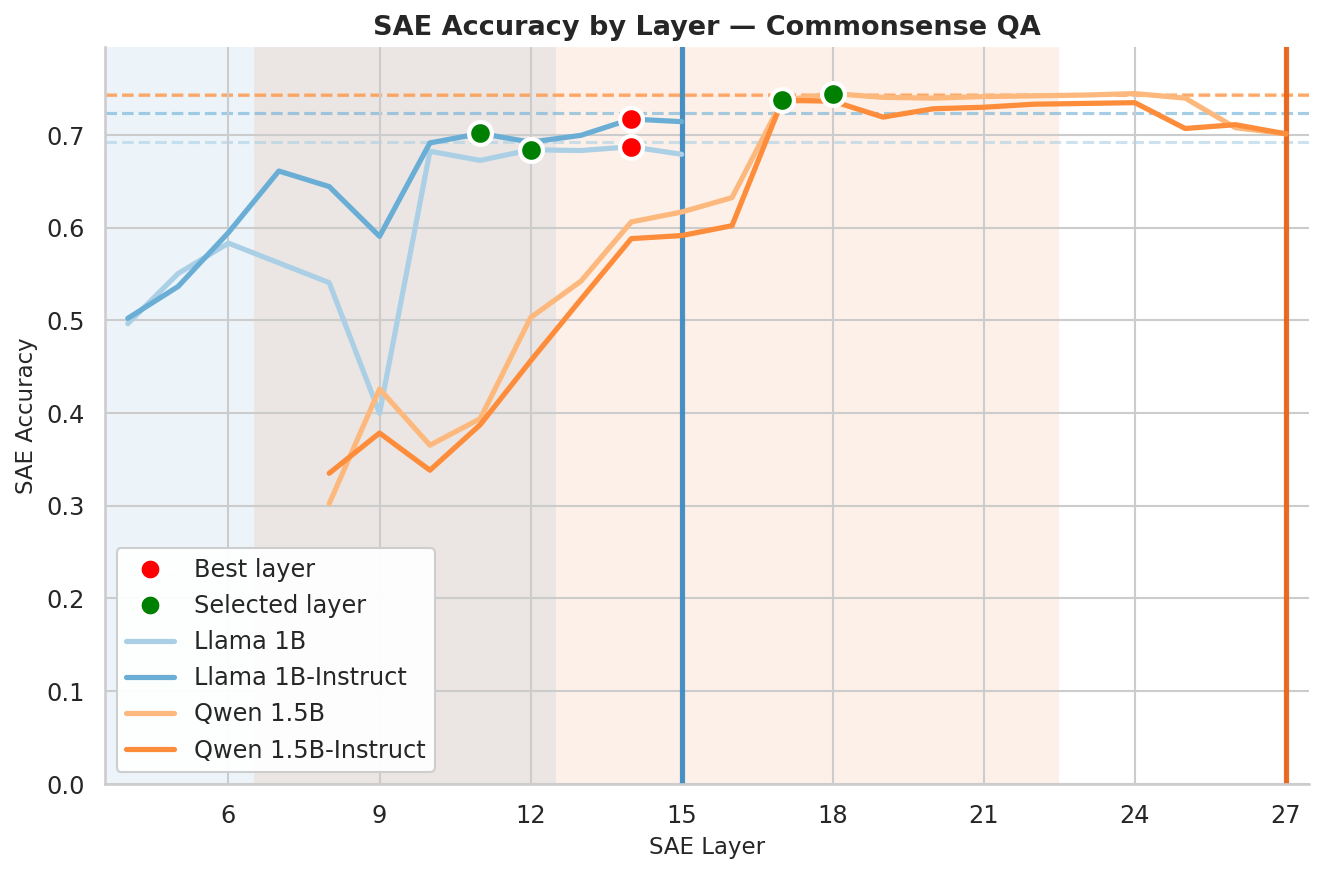}
  \caption{CommonsenseQA.}
  \label{fig:sae-layer-sensitivity-commonsenseqa}
\end{subfigure}
\caption{Accuracy change $\Delta$ versus SAE insertion layer across datasets and SAE configurations.}
\label{fig:sae-layer-sensitivity-all}
\end{figure*}

\paragraph{Takeaways.}
Across all three datasets: (i) there typically exists a contiguous band of intermediate layers where the post-insertion accuracy drop is small; (ii) the optimal insertion layer can shift by several blocks across model families and datasets; and (iii) larger SAEs (higher latent dimension $H$) tend to be more forgiving to insertion, while more aggressive sparsity (smaller $k$) increases layer sensitivity. These trends motivate the layer-selection rule in Section~\ref{sec:ftandsae}: we search within $\mathcal{L}_{\mathrm{mid}}$ to avoid both input-adjacent and head-adjacent regimes, and we report both the selected layer (restricted to $\mathcal{L}_{\mathrm{mid}}$) and the best layer over the full sweep.

\paragraph{Depth trade-off: downstream preservation vs. feature utility.}
To quantify the trade-off between preserving downstream behavior and obtaining useful/interpretable features, we compare necessity test scores at two insertion depths (L10 and L13). For Llama-3.2-1B-Instruct on OpenBookQA, Layer 10 is the selected layer under our middle-half selection policy, while Layer 13 is the best performance layer; both points are marked in Figure~\ref{fig:sae-layer-sensitivity-openbookqa}. Table~\ref{tab:layer-depth-tradeoff} shows that the deeper layer (L13) consistently yields smaller NLL increase for most $k$, indicating worse task behavior preservation, while also exhibiting substantially lower flip rates, indicating weaker controllable feature effects.
This supports the practical choice of intermediate layers for auditing: deeper insertion can preserve accuracy better, but the resulting features tend to be less interpretable/usable for intervention-based analysis.

\begin{table}[t]
\centering
\caption{Trade-off across insertion depth: intermediate insertion layer better preserves downstream behavior  and yields stronger intervention effects with lower $\Delta$NLL and higher flip rate (FR), suggesting higher feature usefulness for interpretability.}
\label{tab:layer-depth-tradeoff}

\begin{tabular}{ccccc}
\toprule
$k$ & $\Delta$NLL (L10) & $\Delta$NLL (L13) & FR (L10) & FR (L13) \\
\midrule
10  & \textbf{0.6498}  & 0.2897  & \textbf{0.384} & 0.182 \\
25  & \textbf{1.3853}  & 0.8533  & \textbf{0.558} & 0.354 \\
50  & \textbf{2.1468}  & 2.1452  & \textbf{0.646} & 0.460 \\
75  & \textbf{4.4718}  & 2.9893  & \textbf{0.836} & 0.582 \\
100 & \textbf{8.2704}  & 4.4470  & \textbf{0.964} & 0.656 \\
150 & \textbf{10.3935} & 9.8442  & \textbf{0.998} & 0.756 \\
200 & 9.4899  & \textbf{10.3416} & \textbf{1.000} & 0.862 \\
\bottomrule
\end{tabular}
\vspace{-2mm}
\end{table}

\subsection{Orthogonality Ablations Across SAE/AE Variants}
\label{app:orth-ablation}

\paragraph{Orthogonality across SAE/AE variants.}
To study how architecture and regularization affect disentanglement, Table~\ref{tab:orthogonality-ablation} compares dense AE (D) and sparse SAE (S) variants across orthogonality metrics and task accuracy (Acc.). 
For the orthogonal-AE baseline, we add a penalty on the off-diagonal mass of the feature Gram matrix of the intermediate dense latent. Let $\hat{Z}\in\mathbb{R}^{N\times d}$ denote minibatch latent activations after centering each feature and normalizing columns to unit $\ell_2$ norm, and let $G=\hat{Z}^{\top}\hat{Z}$. We use
\[
\mathcal{L}_{\mathrm{ortho}}
= \frac{1}{d(d-1)}\sum_{i\neq j}G_{ij}^2
= \frac{1}{d(d-1)}\left\|G-I_d\right\|_{F,\mathrm{off}}^2,
\]
and train with $\mathcal{L}=\mathcal{L}_{\mathrm{task}}+\lambda_{\mathrm{rec}}\mathcal{L}_{\mathrm{rec}}+\lambda_{\mathrm{ortho}}\mathcal{L}_{\mathrm{ortho}}$. In the table, $\mathrm{OW}$ denotes $\lambda_{\mathrm{ortho}}$, and $\mathrm{OW}=\text{--}$ indicates no orthogonality penalty.

Overall, sparse SAEs yield the strongest near-orthogonality (up to $99.10\%$ at $k{=}16384/256$) and lowest off-diagonal correlation magnitudes, while dense AEs improve substantially as OW increases (e.g., better stable rank and lower mean-squared correlation at $\mathrm{OW}=1.0$). Across settings, this indicates an orthogonality--accuracy tradeoff: improving feature orthogonality is generally accompanied by some drop in task accuracy (within roughly $70.8$--$73.4$ in our runs).

\begin{table}[t]
\centering
\caption{Orthogonality comparison across sparse SAE and dense AE variants. Lower is better for mean/squared/Frobenius off-diagonal metrics; higher is better for stable rank and near-orthogonality. G-Abs and G-Sq stand for the mean absolute/squared value of the gram off-diagnoal entries.}
\label{tab:orthogonality-ablation}
\begin{tabular}{lllcccccc}
\toprule
Sparse Setup & Space & OW & G-Abs $\downarrow$ & G-Sq $\downarrow$ & F-norm $\downarrow$ & SR $\uparrow$ & \%$<0.1$ $\uparrow$ & Acc. \\
\midrule
D 512           & Text          & --    & 0.153 & 0.057 & 0.238 & 4.37   & 56.81\% & 72.6 \\
D 2048          & Pre-latent    & --    & 0.227 & 0.083 & 0.289 & 3.76   & 28.90\% & 72.6 \\
D 2048          & Latent        & 0.1   & 0.124 & 0.055 & 0.235 & 4.68   & 65.11\% & 71.2 \\
D 2048          & Latent        & 1.0   & 0.090 & 0.016 & 0.125 & 16.95  & 65.56\% & 70.8 \\
S $2048/256$    & Latent        & --    & 0.095 & 0.028 & 0.167 & 7.79   & 75.43\% & 73.4 \\  
S $16384/256$   & Latent        & --    & 0.008 & 0.002 & 0.047 & 25.02  & 98.67\% & 73.2\\
S $16384/256$   & Latent        & 0.1   & 0.006 & 0.002 & 0.040 & 22.34  & 99.10\% & 72.6 \\
D 16384         & Latent        & 0.1   & 0.100 & 0.029 & 0.169 & 7.61   & 66.41\% & 71.2 \\
D 16384         & Latent        & 1.0   & 0.090 & 0.015 & 0.124 & 22.51  & 65.98\% & 71.4 \\
\bottomrule
\end{tabular}
\end{table}

\subsection{Qualitative Attribution: Linking Predictions to Training Evidence}
\label{app:case-studies}
We now present case studies illustrating how our method links a test prediction to specific training evidence.
For each case, we report the model prediction, the most influential training examples, and token-level heatmaps obtained by projecting representation-level influence back to the input space.

\paragraph{Case study protocol.}
We select test instances that are predicted incorrectly by the pretrained model but become correct after finetuning, so that the attributions reflect task-specific learning rather than generic priors. We run this procedure for both Llama-3.2-1B-Instruct and Qwen2.5-1.5B-Instruct; for readability, the main text shows one representative example and additional cases (including Qwen-based results) are provided in Appendix~\ref{app:case-studies}.

For each selected test example, we compute representation-level influence $\mathrm{IFR}$ and use it to rank influential latent features and influential training examples. We then visualize token-level influence by projecting the influential latent signal onto SAE activations and color tokens by their dominant influential latent, with intensity given by activation $\times$ influence (normalized per sentence). We also interactively support hovering over each latents to see its corresponding tokens.

\paragraph{Case Study B: Openbook Influence retrieval identifies shared physical mechanisms.}
\label{app:caseB}

This example illustrates that RepInfLLM retrieves evidence at the level of physical mechanisms rather than exact answer strings. The test question asks which object can serve as an electrical conductor, with the correct answer “a penny.” The most influential training sample instead asks what flows when one electrical conductor contacts another electrical conductor, but it activates the same underlying concept: conductors permit the flow of electricity or power. The highlighted latent features concentrate on “one electrical conductor,” “another electrical conductor,” “flow,” and “power,” showing that the influence signal is localized to the mechanism-bearing phrase rather than to superficial option overlap. This suggests that representation-level influence can recover training evidence that supports the test prediction through shared scientific structure.

\begin{figure}[t]
\centering
\begin{subfigure}[t]{0.8\linewidth}
  \centering
  \includegraphics[width=\linewidth]{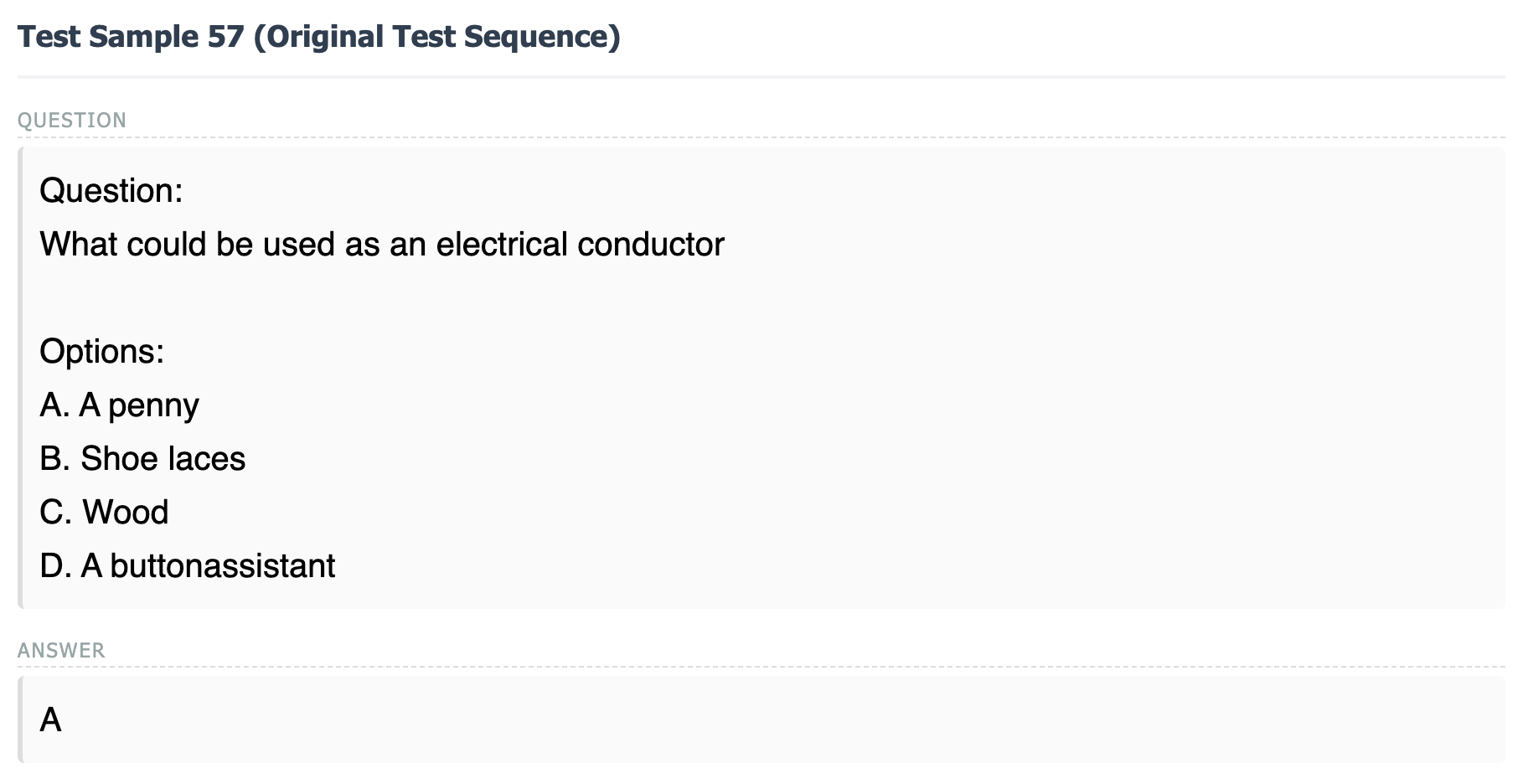}
  \caption{Test input (token-level influence).}
\end{subfigure}

\vspace{2mm}

\begin{subfigure}[t]{0.8\linewidth}
  \centering
  \includegraphics[width=\linewidth]{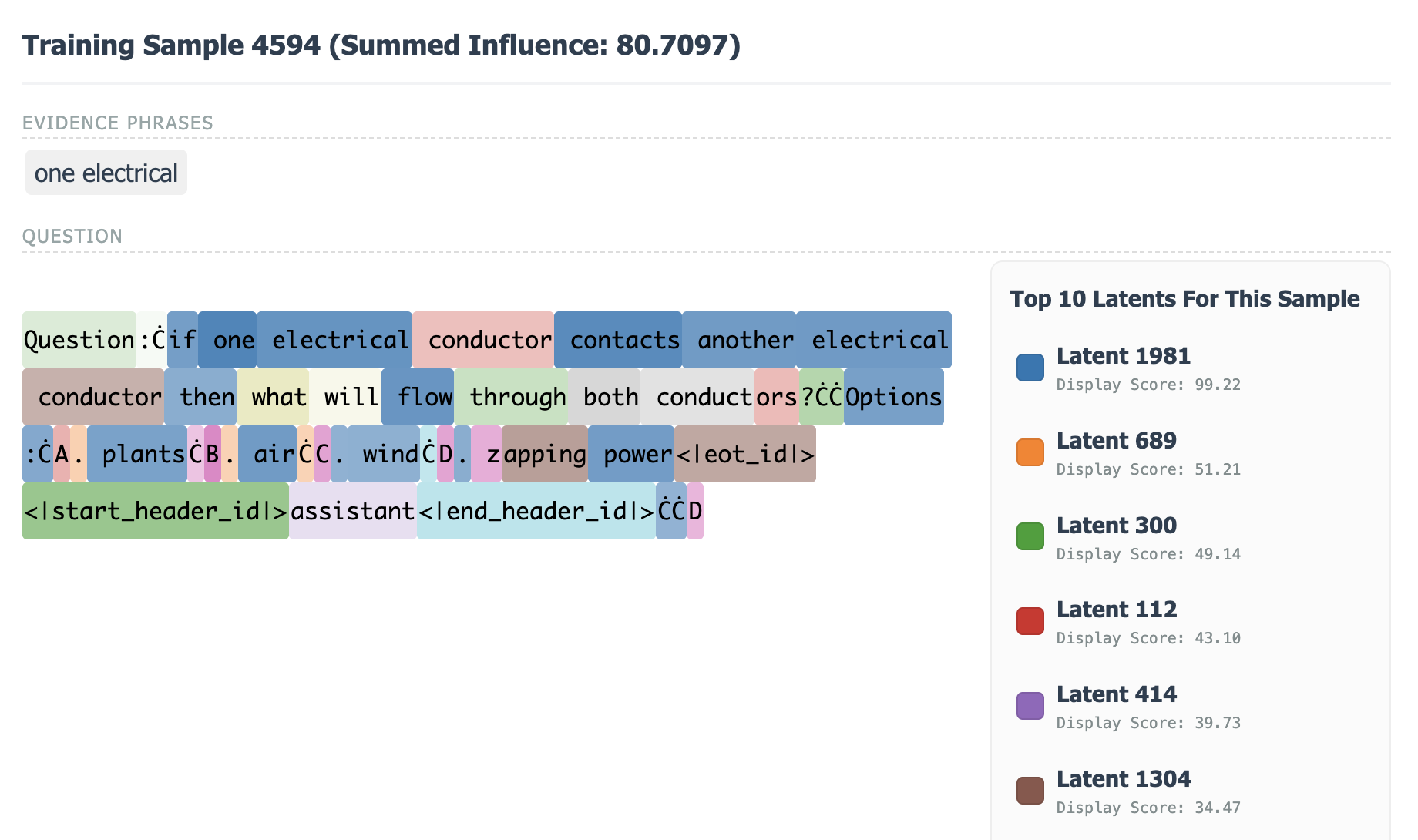}
  \caption{Most influential training example (token-level influence).}
\end{subfigure}
\caption{Case Study B (Llama-3.2-1B-Instruct): Electrical conductor.}
\label{fig:heatmap-case2}
\end{figure}

\paragraph{Case Study C: MedQA influence highlight Asthma exacerbation and small-airway obstruction}
\label{app:heatmap-case3}

\begin{figure}[t]
\centering
\begin{subfigure}[t]{0.8\linewidth}
  \centering
  \includegraphics[width=\linewidth]{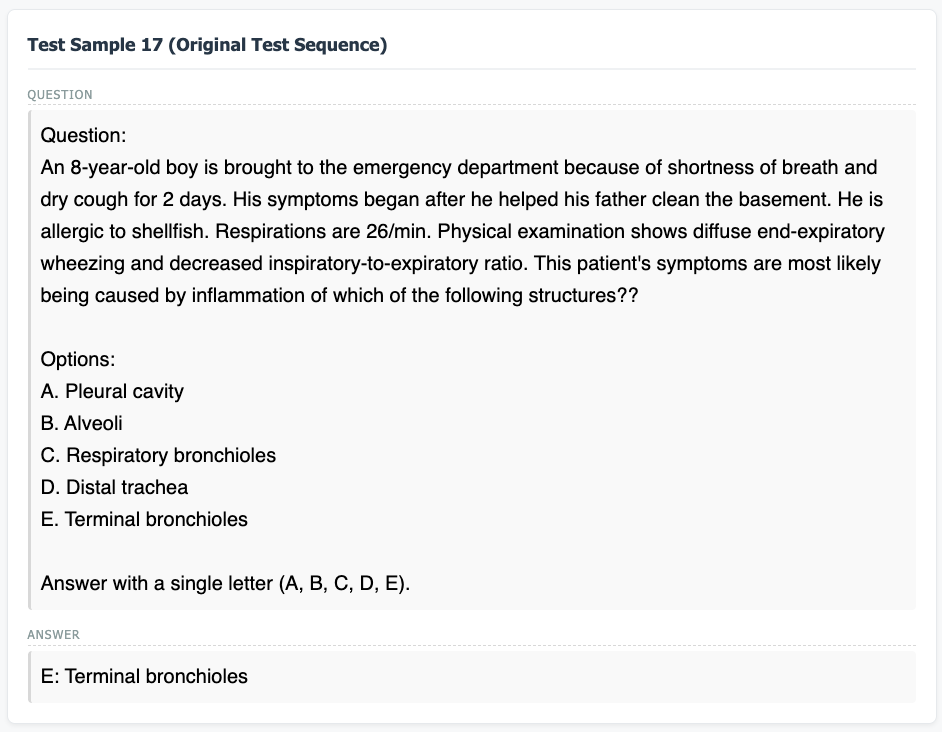}
  \caption{Test input (token-level influence).}
\end{subfigure}

\vspace{2mm}

\begin{subfigure}[t]{0.8\linewidth}
  \centering
  \includegraphics[width=\linewidth]{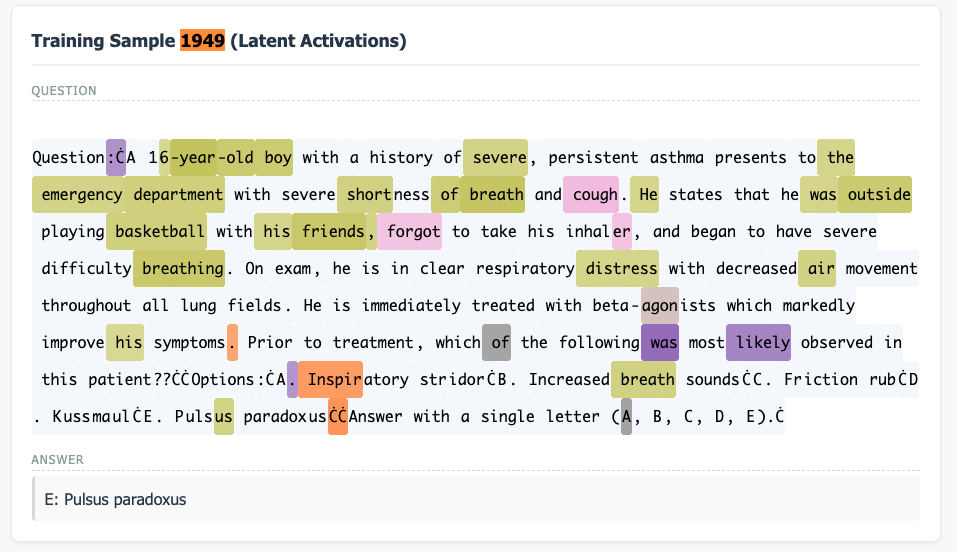}
  \caption{Most influential training example (token-level influence).}
\end{subfigure}
\caption{Case Study C (Qwen2.5-1.5B-Instruct): asthma exacerbation and small-airway obstruction.}
\label{fig:heatmap-case3}
\end{figure}

The test example in Figure~\ref{fig:heatmap-case3} describes an 8-year-old with shortness of breath and dry cough after environmental exposure, diffuse end-expiratory wheezing, and a decreased inspiratory-to-expiratory ratio. The correct choice is \emph{terminal bronchioles}, consistent with inflammation and narrowing of small airways in an obstructive process.

The retrieved training example concerns a severe asthma presentation and asks for a classic physiologic finding (pulsus paradoxus). Although the questions differ, influential latents activate on shared features related to bronchospasm and respiratory distress (e.g., wheeze, reduced airflow, difficulty breathing). This illustrates that representation-level influence can connect test predictions to training evidence through a shared obstruction motif, without requiring exact surface-form matching.

\paragraph{Case Study D: Reasoning augmentation sharpens deep-sea animal influence.}
\label{app:heatmap-case4}
For this case study, we did something different with the setup. We construct a reasoning-augmented training sample by keeping the original question answer pair fixed and appending a short rationale that makes the relevant semantic relation explicit. We then recompute representation level influence and compare it with the baseline heatmap to test whether the added reasoning sharpens attribution from surface overlap toward concept level evidence.

\begin{figure}[t]
\centering
\begin{subfigure}[t]{\linewidth}
  \centering
  \includegraphics[width=0.8\linewidth]{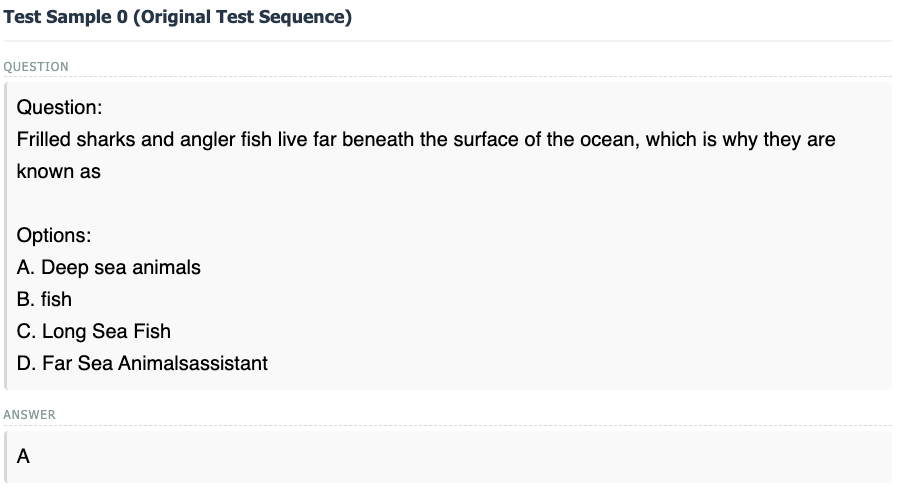}
  \caption{Test input (token-level influence).}
\end{subfigure}
\begin{subfigure}[t]{\linewidth}
  \centering
  \includegraphics[width=0.8\linewidth]{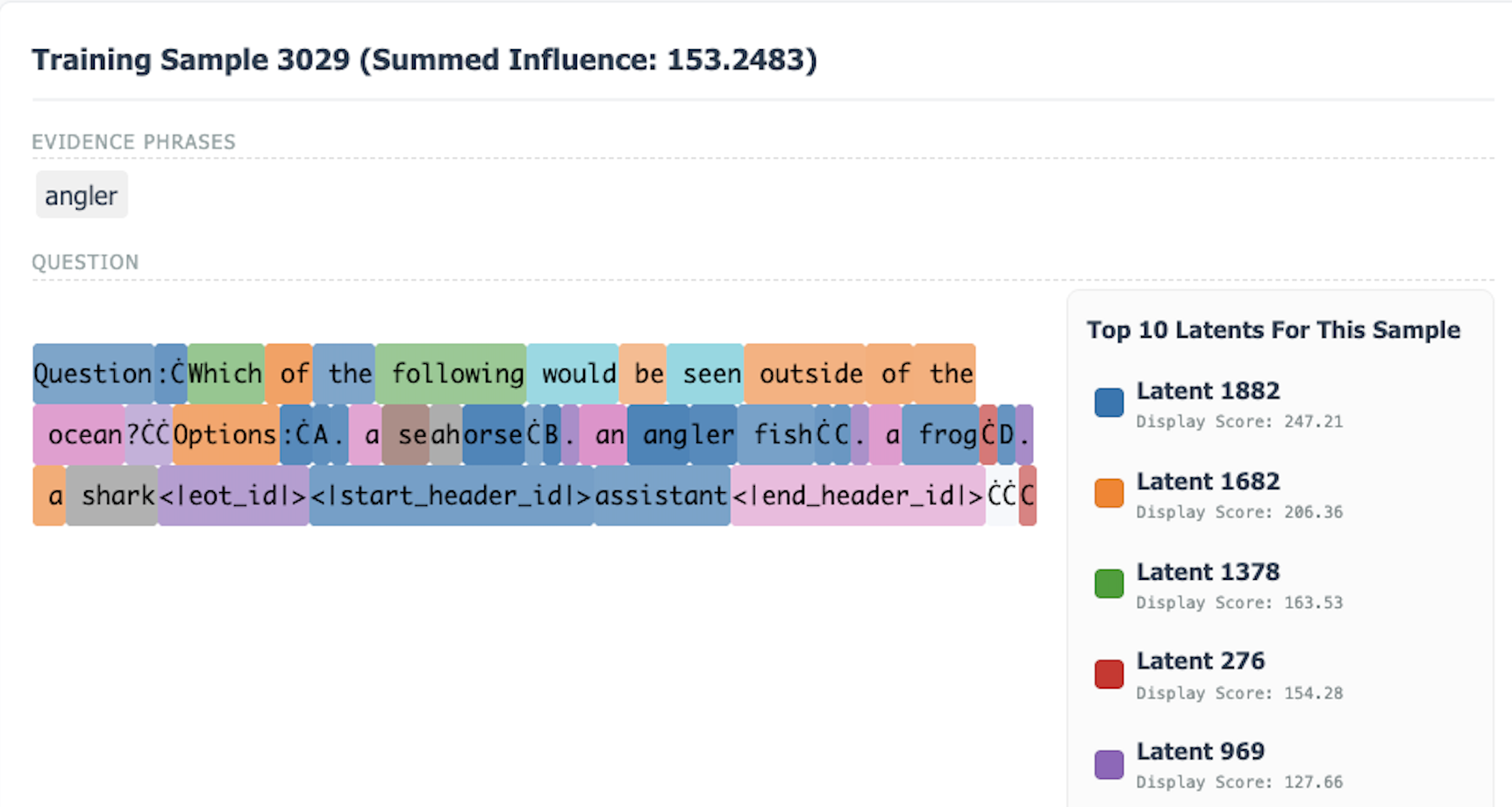}
  \caption{Most influential training example (token-level influence).}
\end{subfigure}
\begin{subfigure}[t]{\linewidth}
  \centering
  \includegraphics[width=0.8\linewidth]{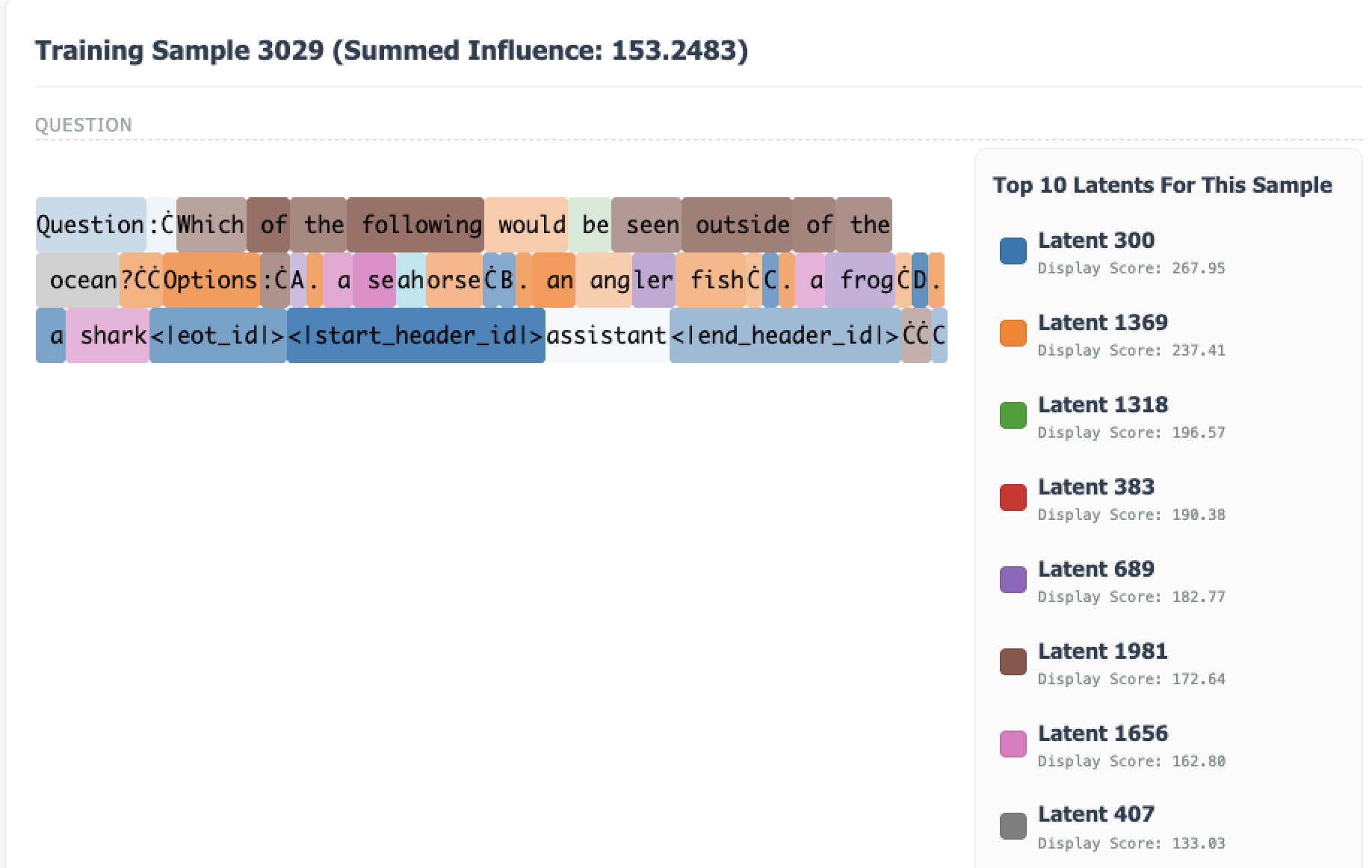}
  \caption{Most influential training example with reasoning augmentation.}
\end{subfigure}
\caption{Case Study D. Reasoning augmented heatmap comparison}
\label{fig:heatmap-case4}
\end{figure}

Figure~\ref{fig:heatmap-case4} shows how adding an explicit reasoning cue changes the retrieved evidence. The test question asks why frilled sharks and angler fish are classified as deep-sea animals, requiring the model to connect these species to organisms living far below the ocean surface. In the baseline heatmap, the retrieved training example is relevant mainly because it contains ``angler fish'' among the answer choices, but the attribution remains diffuse across the multiple-choice sequence and does not clearly isolate the required reasoning. After adding the reasoning chain, the influence becomes sharper: highlighted regions concentrate on the phrase ``outside of the ocean'' and the animal tokens, especially ``angler fish'' and ``shark.'' This suggests that reasoning augmentation helps the latent features align animal identity with ocean depth context, yielding a clearer link between the retrieved training sample and the test answer.

\section{Experimental Details}
\label{app:exp-details}

This appendix records the settings needed to reproduce the experiments in Section~\ref{sec:exp}.



\subsection{Models and fine-tuning}
\label{app:models-ft}

\begin{itemize}
    \item \textbf{Base models.} Qwen2.5-1.5B-Instruct and Llama-3.2-1B-Instruct.
    \item \textbf{Fine-tuning.} We experiment with both full fine-tuning and LoRA. LoRA is faster and more memory-efficient; in our setting it also yields better task performance, consistent with reduced catastrophic forgetting under constrained compute.
    \item \textbf{Optimization.} AdamW with learning rate $2\times10^{-4}$, batch size 32, 4 epochs, maximum sequence length 512..
    \item \textbf{Evaluation.} Accuracy over multiple-choice options $\{A,B,C,D,E\}$, extracted from the model generation using a deterministic option parser.
\end{itemize}

\subsection{SAE training}
\label{app:sae-training}

\begin{itemize}
    \item \textbf{Insertion layers.} Llama: layers 4--12; Qwen: layers 7-22. The reported ``selected layer'' is the best-performing layer within $\mathcal{L}_{\mathrm{mid}}$.
    \item \textbf{Objective.} Reconstruction MSE on the instrumented hidden state, optionally trained jointly with the downstream task loss when enabled (see Table~\ref{tab:exp-hparams-clean}).
    \item \textbf{SAE size and sparsity.} Latent dimension $H\in\{16384,2048,512\}$ and Top-$k$ sparsity $k\in\{256,128,64\}$.
    \item \textbf{Training data and steps.} \texttt{2 epochs}.
\end{itemize}

\paragraph{Latent size selection.}
The alternative visualization uses the task-sized SAE with latent dimension $H=2048$ and Top-$k=256$, whereas some exploratory visualizations use larger SAEs to inspect finer-grained concepts. We choose the compact SAE for the 1B-scale OpenbookQA experiments because the finetuning set contains only $\sim$5k examples and exact influence computation scales with the number of latent coordinates. The smaller latent space produces a denser heatmap with densor activation, making full influence sweeps and heatmap rendering readable.

\subsection{Influence computation and filtering}
\label{app:influence-details}

\paragraph{Gradient pre-filtering.}
\label{app:filtering}
We rank training examples by cosine similarity between their gradients and the test gradient:
\begin{equation}
\begin{aligned}
\mathrm{sim}(z_i, z_{\text{test}})
&= \frac{\langle g_i, g_{\text{test}} \rangle}{\|g_i\|\,\|g_{\text{test}}\|}, \\
g_i &= \nabla_{\theta}\ell(h_{\theta}(x_i), y_i), \\
g_{\text{test}} &= \nabla_{\theta}\ell(h_{\theta}(x_{\text{test}}), y_{\text{test}}).
\end{aligned}
\end{equation}
We retain the top \{1\% / 5\% / 10\%\} candidates for exact influence computation.

\paragraph{iHVP approximation.}
We compute $H_{\theta_2}^{-1}g_{\text{test}}$ using conjugate gradient (CG), with damping \texttt{1e-3} and iteration budget \texttt{20}.
Unless otherwise stated, the empirical Hessian is formed using a batched loss with batch size 8 to improve curvature stability.

\paragraph{Influence target.}
Unless otherwise stated, influence is computed on the test loss.

\begin{table}[t]
\centering
\caption{Key experimental hyperparameters used throughout the appendix.}
\label{tab:exp-hparams-clean}

\begin{tabular}{p{0.22\linewidth} p{0.72\linewidth}}
\toprule
\textbf{Category} & \textbf{Setting} \\
\midrule
Fine-tuning & LoRA; AdamW, lr $2\times10^{-4}$, batch size 32, 4 epochs, max length 512. \\
SAE & Layer sweep: Llama 4--15, Qwen 4--26; objective: reconstruction MSE; $H\in\{2048,1024,512\}$, $k\in\{256,128,64\}$. \\
Filtering & Gradient similarity pre-filtering; retained fraction 1\%--10\%. \\
Influence & iHVP via CG using batched empirical loss (batch size 8); damping $10^{-3}$; \texttt{20} iterations. \\
\bottomrule
\end{tabular}
\vspace{-2mm}
\end{table}


\section{Scaling to 1B-Parameter LLMs}
\label{app:scaling_1b}

Scaling representation influence to 1B-parameter LLMs is challenging due to: (i) memory blow-ups from naïvely materializing second-order objects or large JVP tensors; and (ii) numerical instability when the empirical Hessian exhibits negative curvature, which can break iterative solvers.

\subsection{Memory scaling via contraction order}
\label{app:memory-scaling}

We compute the representation influence for latent dimension $j$ as
\begin{align}
\mathcal{I}_j^r(z^r_{\text{train}}, z_{\text{test}})
&= - \underbrace{g_{\text{test}}^\top H_{\theta_2}^{-1}}_{\text{iHVP term}}
\;\underbrace{\mathrm{JVP}(G, r_{\text{train}}, e_j)}_{\text{per-latent JVP}},
\label{eq:repr_infl_app}
\end{align}
where $H_{\theta_2}$ is the Hessian of the training objective restricted to downstream parameters $\theta_2$ and $g_{\text{test}}$ is the test gradient.

\paragraph{Avoiding Hessian materialization.}
We never explicitly construct $H_{\theta_2}$. Instead, we rely on Hessian-vector products (HVPs) computed via automatic differentiation, enabling CG solves without storing the Hessian.

\paragraph{Avoiding Jacobian materialization.}
A naive implementation computes all $\{\mathrm{JVP}(G,r_{\text{train}},e_j)\}_{j=1}^{H}$ in one autograd call by concatenating basis vectors; this leads to catastrophic memory usage. Profiling shows the per-feature JVP stage can allocate $\sim$10--15\,GB on GPT-2, and aggregation can add another $\sim$10--15\,GB, pushing peak memory toward $\sim$40\,GB. Dense offloading is worse: storing the full Jacobian scales as $\mathcal{O}(H\cdot |\theta_2|)$ and can reach hundreds of GB at 1B scale.

We therefore compute influence in a \emph{streamed contraction} manner:
\begin{enumerate}[leftmargin=*, itemsep=1pt, topsep=2pt]
\item \textbf{Precompute iHVP once per test input.} Compute $s_{\text{test}}^\top \triangleq g_{\text{test}}^\top H_{\theta_2}^{-1}$ once and reuse it across all latent dimensions.
\item \textbf{Fuse JVP with contraction.} For each $j$, compute $r_j \triangleq \mathrm{JVP}(G,r_{\text{train}},e_j)$ and immediately contract
$\mathcal{I}_j^r = - s_{\text{test}}^\top r_j$, then discard $r_j$.
\end{enumerate}
This reduces storage from $\mathcal{O}(H\cdot |\theta_2|)$ to $\mathcal{O}(|\theta_2|)$ and avoids Jacobian storage.
In our Llama-3.2-1B profiling, this reduces the peak from an estimated dense-materialization regime ($\sim$400\,GB) to a practical tens-of-GB regime.

\subsection{Stabilizing iHVP under noisy curvature}
\label{app:ihvp-stability}

We obtain $s_{\text{test}}^\top$ by solving $H_{\theta_2} v = g_{\text{test}}$ using CG, where $Hv$ is computed via autograd HVPs.
Although CG assumes symmetric positive definite curvature, too-small batches yield noisy empirical Hessians with occasional negative curvature, manifested as $p^\top Hp<0$ and early termination or degenerate solutions.
Empirically, batch size $\le 4$ can cause a large fraction of influence entries to collapse to near-zero due to solver instability.
We therefore define $H_{\theta_2}$ using a batched empirical loss with batch size 8 for stable curvature in all 1B runs.

\subsection{Compute-time scaling and accelerations}
\label{app:compute-scaling}

Compute is dominated by the latent loop if we explicitly sweep per-feature JVPs. Without further reductions, on Llama-3.2-1B the per-$(z_{\ttrain}, z_{\ttest})$ cost decomposes into roughly $\sim$1\,s (forward/backward), $\sim$5\,s (CG iHVP), and $\sim$30--60\,s (full JVP sweep, depending on model and SAE size), which is infeasible when repeated over many training candidates.

\begin{table}[t]
\centering
\caption{Runtime and memory breakdown for 1B-scale influence computation per $(z_{\ttrain}, z_{\ttest})$ pair, based on our profiling. ``JVP sweep'' corresponds to streamed contraction (no Jacobian storage).}
\label{tab:scaling-llama1b}

\begin{tabular}{lcc}
\toprule
Component & Time (s) & Peak GPU mem \\
\midrule
Forward/backward & $\sim$1 & -- \\
CG iHVP solve & $\sim$5 & -- \\
Llama-3.2-1B JVP sweep & $\sim$35 & $\sim$50\,GB \\
Qwen-2.5-1.5B JVP sweep & $\sim$60 & $\sim$100\,GB \\
\bottomrule
\end{tabular}
\vspace{-4mm}
\end{table}

We apply two practical reductions:
\begin{enumerate}[leftmargin=*, itemsep=1pt, topsep=2pt]
\item \textbf{Task-sized SAE.} Since MedQA fine-tuning has $\sim$9k training samples, we train effective SAEs with latent size $H=512$ and Top-$k=64$, reducing the JVP sweep to $\sim$30\,s and total to $\sim$40\,s per candidate training example.
\item \textbf{Gradient-similarity pre-filtering.} We compute exact influence only on the top $1\%$--$10\%$ training candidates ranked by gradient cosine similarity.
\end{enumerate}
With these reductions, end-to-end influence computation is $\sim$1 hour per test sample in our setting.

\subsection{Further acceleration via the gradient-derivative formulation (Sec.~3.4)}
\label{app:grad-derivative-accel}

Beyond streamed JVP contraction, we exploit the gradient-derivative reformulation from Sec.~3.4, which eliminates the explicit per-feature JVP sweep.
Recall that representation influence can be written as
\begin{align}
\mathcal{I}^r(z^r_{\text{train}}, z_{\text{test}})
= - s_{\text{test}}^\top G(r_{\text{train}}),
\end{align}
where $s_{\text{test}}^\top \triangleq g_{\text{test}}^\top H_{\theta_2}^{-1}$ and $G(r_{\text{train}})=\nabla_{\theta_2}\ell(h_{\theta_2}(r_{\text{train}}),y_{\text{train}})$.
Instead of iterating over $j$ and computing $\mathrm{JVP}(G,r_{\text{train}},e_j)$, we directly differentiate the scalar projection with respect to $r_{\text{train}}$:
\begin{align}
\nabla_{r_{\text{train}}}\big(s_{\text{test}}^\top G(r_{\text{train}})\big).
\end{align}
This single reverse-mode pass produces the full latent-level influence vector $\{\mathcal{I}_j^r\}_{j=1}^{H}$, replacing an $\mathcal{O}(H)$ JVP loop with one sensitivity backpropagation.

\begin{table}[t]
\centering
\caption{Runtime breakdown (seconds) per $(z_{\ttrain}, z_{\ttest})$ pair using the gradient-derivative formulation on Qwen-2.5-1.5B.}
\label{tab:qwen15b_runtime}
\vspace{-1mm}
\begin{tabular}{lc}
\toprule
Operation & Time (s) \\
\midrule
Train forward & 0.2388 \\
Compute batched iHVP & 3.4927 \\
Gradient projection & 0.2228 \\
Sensitivity backprop & 0.4834 \\
Attribution \& aggregation & 0.0001 \\
\midrule
\textbf{Total} & \textbf{4.4378} \\
\bottomrule
\end{tabular}
\vspace{-3mm}
\end{table}

\begin{table}[t]
\centering
\caption{Runtime breakdown (seconds) per $(z_{\ttrain}, z_{\ttest})$ pair using the gradient-derivative formulation on Llama-3.2-1B.}
\label{tab:llama1b_runtime}
\vspace{-1mm}
\begin{tabular}{lc}
\toprule
Operation & Time (s) \\
\midrule
Train forward & 0.0947 \\
Compute batched iHVP & 1.6803 \\
Gradient projection & 0.1443 \\
Sensitivity backprop & 0.2460 \\
Attribution \& aggregation & 0.0001 \\
\midrule
\textbf{Total} & \textbf{2.1654} \\
\bottomrule
\end{tabular}
\vspace{-3mm}
\end{table}

\paragraph{Overall compute comparison.}
Table~\ref{tab:compute_comparison} compares the streamed-JVP pipeline with the gradient-derivative formulation.
The latter removes the $\sim$30--60\,s JVP sweep (Table~\ref{tab:scaling-llama1b}) and reduces total per-pair runtime to $\approx 2$--$4$ seconds, corresponding to a $\sim 10\times$--$20\times$ practical speedup.

\begin{table}[t]
\centering
\caption{End-to-end per-$(z_{\ttrain}, z_{\ttest})$ runtime comparison (seconds).}
\label{tab:compute_comparison}
\begin{tabular}{lcc}
\toprule
Model & Streamed JVP & Gradient-Derivative \\
\midrule
Llama-3.2-1B & $\sim$40 & 2.17 \\
Qwen-2.5-1.5B & $\sim$60 & 4.44 \\
\bottomrule
\end{tabular}
\end{table}

Combined with batched iHVP solves and gradient-similarity pre-filtering, this reduction makes representation-level influence computation practical for 1B--1.5B parameter LLMs under commodity multi-GPU constraints.

\input{_sa_full_proof}

%% file: _sa_full_proof.tex
\section{Derivations}
\subsection{Exact Path-Integral Decomposition}
\label{app:path_integral_decomposition}

Let $G:\mathbb{R}^{d_l}\rightarrow\mathbb{R}^{|\theta_2|}$ be continuously differentiable.
Fix $r_0, r_{\text{train}} \in \mathbb{R}^{d_l}$ and define the straight-line path
\begin{equation}
r(\alpha)=r_0+\alpha(r_{\text{train}}-r_0),
\qquad \alpha\in[0,1].
\end{equation}

\paragraph{Path integral identity.}
By the chain rule,
\begin{equation}
\frac{d}{d\alpha} G(r(\alpha))
=
J_G(r(\alpha))\,\frac{dr(\alpha)}{d\alpha}
=
J_G(r(\alpha))(r_{\text{train}}-r_0),
\end{equation}
where $J_G(r)$ denotes the Jacobian of $G$.
Applying the fundamental theorem of calculus,
\begin{equation}
\begin{aligned}
G(r_{\text{train}})-G(r_0)
&= \int_0^1
\frac{d}{d\alpha} G(r(\alpha))\\
&=
\int_0^1
J_G(r(\alpha))(r_{\text{train}}-r_0)\,d\alpha.
\label{eq:path_integral_appendix}
\end{aligned}
\end{equation}

\paragraph{Coordinate decomposition.}
Using the canonical basis expansion
\begin{equation}
r_{\text{train}}-r_0
=
\sum_{j=1}^{d_l}
(r_{\text{train}}^{(j)}-r_0^{(j)})e_j,
\end{equation}
and linearity of integration,
\begin{align}
G(r_{\text{train}})
&=
G(r_0)
+
\sum_{j=1}^{d_l}
(r_{\text{train}}^{(j)}-r_0^{(j)})
\left(
\int_0^1
J_G(r(\alpha))e_j\,d\alpha
\right).
\label{eq:latent_decomposition_appendix}
\end{align}

Equations~\eqref{eq:path_integral_appendix}--\eqref{eq:latent_decomposition_appendix}
hold exactly under the sole assumption that $G$ is continuously differentiable.

\subsection{Derivative Swapping with activation}
\label{app:derivative_swap_proof}
While the Jacobian-vector product (JVP) formulation in \eqref{eq:per_representation_if_jvp} provides a rigorous attribution mechanism, computing it naively is computationally prohibitive. Evaluating the JVP for each latent dimension $j \in \{1, \dots, d_l\}$ requires $d_l$ separate forward-mode passes. For a sparse autoencoder with thousands of features, this $\mathcal{O}(d_l)$ complexity leads to severe computational bottlenecks and memory allocation limits.

We can achieve a massive optimization by exploiting the structure of the computation graph and the symmetry of mixed partial derivatives (Clairaut's Theorem). We transition from evaluating $d_l$ directional derivatives to performing a single reverse-mode gradient pass. 

\vspace{2pt}
\noindent\textbf{The Independence of Upstream Activations.\;\;}
Recall that the model is split at layer $l$, such that the latent representation $r_\ttrain = h_{\theta_1}(x_\ttrain)$ is produced entirely by upstream parameters $\theta_1$, while we differentiate with respect to the downstream parameters $\theta_2 = \{\theta:\theta_{>l}\}$. Because $r_\ttrain$ serves as an input to the downstream computation and does not depend on $\theta_2$, it is treated as a constant with respect to $\theta_2$. Therefore, the parameter gradient of the activation is strictly zero:
\begin{equation}\label{eq:app_graph_cut}
\frac{\partial r_\ttrain^{(j)}}{\partial \theta_2} = \mathbf{0}.
\end{equation}

\vspace{2pt}
\noindent\textbf{Equivalence of Inside and Outside Weighting.\;\;}
Let $\Delta^{(j)} = \frac{\partial \ell}{\partial r_\ttrain^{(j)}}$ denote the sensitivity of the training loss to the $j$-th latent feature. In our operationalized code, we define a feature's loss contribution as $s^{(j)} = r_\ttrain^{(j)} \Delta^{(j)}$. 
Taking the gradient of this contribution with respect to the downstream parameters $\theta_2$, we apply the product rule:
\begin{equation}
\frac{\partial s^{(j)}}{\partial \theta_2} 
\;=\; 
\frac{\partial}{\partial \theta_2} \left( r_\ttrain^{(j)} \Delta^{(j)} \right) 
\;=\; 
\underbrace{\frac{\partial r_\ttrain^{(j)}}{\partial \theta_2}}_{=\mathbf{0}} \Delta^{(j)} 
+ 
r_\ttrain^{(j)} \frac{\partial \Delta^{(j)}}{\partial \theta_2}.
\end{equation}
Because the first term vanishes due to the computation graph cut \eqref{eq:app_graph_cut}, we are left with:
\begin{equation}
\frac{\partial s^{(j)}}{\partial \theta_2} 
\;=\; 
r_\ttrain^{(j)} \frac{\partial}{\partial \theta_2} \left( \frac{\partial \ell}{\partial r_\ttrain^{(j)}} \right)
\;=\;
r_\ttrain^{(j)} \frac{\partial G(r_\ttrain)}{\partial r_\ttrain^{(j)}},
\end{equation}
where $G(r_\ttrain) = \nabla_{\theta_2}\ell$. This proves that multiplying the activation ``inside'' the gradient tracker is mathematically identical to multiplying by the activation ``outside'' the mixed-partial term, perfectly aligning the empirical implementation with the theoretical Taylor decomposition from Definition~\ref{def:neuron-if}.

\vspace{2pt}
\noindent\textbf{Order Swapping for Constant-Time Evaluation.\;\;}
We now apply the influence contraction with the inverse-HVP test signal $v = H_{\theta_2}^{-1} g_\ttest$. The influence for neuron $j$ becomes:
\begin{equation}\label{eq:swapping_step1}
\mathcal{I}_j^r(z^r_\ttrain, z_\ttest) 
= - \left( \frac{\partial s^{(j)}}{\partial \theta_2} \right)^\top v \\
= - \, r_\ttrain^{(j)} 
\left( 
\frac{\partial}{\partial \theta_2} 
\frac{\partial \ell}{\partial r_\ttrain^{(j)}} 
\right)^\top v .
\end{equation}

By the symmetry of mixed partial derivatives (assuming standard regularity conditions on the loss surface), we can swap the order of differentiation:
\begin{equation}\label{eq:app_mixed_partial_swap}
\left( \frac{\partial}{\partial \theta_2} \frac{\partial \ell}{\partial r_\ttrain^{(j)}} \right)^\top v 
\;=\; 
\frac{\partial}{\partial r_\ttrain^{(j)}} \left( \left( \frac{\partial \ell}{\partial \theta_2} \right)^\top v \right) \\
\;=\; 
\frac{\partial}{\partial r_\ttrain^{(j)}} \Big( G(r_\ttrain)^\top v \Big).
\end{equation}

Define the scalar projection $P = G(r_\ttrain)^\top v$. This scalar represents the alignment between the training parameter gradients and the test signal.  \eqref{eq:app_mixed_partial_swap} reveals that the sensitivity for neuron $j$ is exactly the partial derivative of $P$ with respect to $r_\ttrain^{(j)}$. 

Consequently, the entire vector of sensitivities for all $d_l$ neurons can be computed simultaneously by taking a single gradient of the scalar $P$ with respect to the latent representation $r_\ttrain$:
\begin{equation}\label{eq:app_scalar_projection_grad}
\text{Sensitivity Vector} = \nabla_{r_\ttrain} P = \nabla_{r_\ttrain} \Big( G(r_\ttrain)^\top v \Big).
\end{equation}

Combining \eqref{eq:swapping_step1}  and \eqref{eq:app_scalar_projection_grad}, the final influence vector for all latent features is obtained via an element-wise product (Hadamard product, denoted by $\odot$) with the realized activations:
\begin{equation}\label{eq:app_final_vectorized_influence}
\vec{\mathcal{I}}^r(z^r_\ttrain, z_\ttest) = - \Big( \nabla_{r_\ttrain} P \Big) \odot r_\ttrain.
\end{equation}

This formulation requires only \emph{two} backward passes total: one to construct the computation graph for $G(r_\ttrain)$, and a second to compute the gradient of the scalar projection $P$ with respect to $r_\ttrain$. The time complexity with respect to the feature dimension drops from $\mathcal{O}(d_l)$ to $\mathcal{O}(1)$. In practice, this 
JVP approach enables the simultaneous computation of influences across all latent features and batch samples, allowing our method to gracefully scale to 1B-parameter models and SAEs with tens of thousands of features without materializing explosive Jacobians.

%% file: example_paper.bib
@inproceedings{kingma2013auto,
  title={Auto-encoding variational bayes},
  author={Kingma, Diederik P and Welling, Max},
  booktitle={International Conference on Learning Representations},
  year={2014}
}

@article{hyvarinen2013independent,
  title={Independent component analysis: recent advances},
  author={Hyv{\"a}rinen, Aapo},
  journal={Philosophical Transactions of the Royal Society A: Mathematical, Physical and Engineering Sciences},
  volume={371},
  number={1984},
  pages={20110534},
  year={2013},
  publisher={The Royal Society Publishing}
}

@article{jolliffe2016principal,
  title={Principal component analysis: a review and recent developments},
  author={Jolliffe, Ian T and Cadima, Jorge},
  journal={Philosophical transactions of the royal society A: Mathematical, Physical and Engineering Sciences},
  volume={374},
  number={2065},
  pages={20150202},
  year={2016},
  publisher={the Royal Society publishing}
}

@article{yin2024direct,
  title={Direct preference optimization using sparse feature-level constraints},
  author={Yin, Qingyu and Leong, Chak Tou and Zhang, Hongbo and Zhu, Minjun and Yan, Hanqi and Zhang, Qiang and He, Yulan and Li, Wenjie and Wang, Jun and Zhang, Yue and others},
  journal={arXiv preprint arXiv:2411.07618},
  year={2024}
}

@inproceedings{koh2017understanding,
  title={Understanding black-box predictions via influence functions},
  author={Koh, Pang Wei and Liang, Percy},
  booktitle={International conference on machine learning},
  pages={1885--1894},
  year={2017},
  organization={PMLR}
}

@article{hampel1974influence,
  title={The influence curve and its role in robust estimation},
  author={Hampel, Frank R},
  journal={Journal of the american statistical association},
  volume={69},
  number={346},
  pages={383--393},
  year={1974},
  publisher={Taylor \& Francis}
}

@article{singhal2023large,
  title={Large language models encode clinical knowledge},
  author={Singhal, Karan and Azizi, Shekoofeh and Tu, Tao and Mahdavi, S Sara and Wei, Jason and Chung, Hyung Won and Scales, Nathan and Tanwani, Ajay and Cole-Lewis, Heather and Pfohl, Stephen and others},
  journal={Nature},
  volume={620},
  number={7972},
  pages={172--180},
  year={2023},
  publisher={Nature Publishing Group}
}

@article{grosse2023studying,
  title={Studying large language model generalization with influence functions},
  author={Grosse, Roger and Bae, Juhan and Anil, Cem and Elhage, Nelson and Tamkin, Alex and Tajdini, Amirhossein and Steiner, Benoit and Li, Dustin and Durmus, Esin and Perez, Ethan and others},
  journal={arXiv preprint arXiv:2308.03296},
  year={2023}
}

@article{cunningham2023sparse,
  title={Sparse autoencoders find highly interpretable features in language models},
  author={Cunningham, Hoagy and Ewart, Aidan and Riggs, Logan and Huben, Robert and Sharkey, Lee},
  journal={arXiv preprint arXiv:2309.08600},
  year={2023}
}

@article{marks2024sparse,
  title={Sparse feature circuits: Discovering and editing interpretable causal graphs in language models},
  author={Marks, Samuel and Rager, Can and Michaud, Eric J and Belinkov, Yonatan and Bau, David and Mueller, Aaron},
  journal={arXiv preprint arXiv:2403.19647},
  year={2024}
}

@misc{barez2025chain,
  author = {Barez, Fazl and Wu, Tung-Yu and Arcuschin, Iván and Lan, Michael and Wang, Vincent and Siegel, Noah and Collignon, Nicolas and Neo, Clement and Lee, Isabelle and Paren, Alasdair and Bibi, Adel and Trager, Robert and Fornasiere, Damiano and Yan, John and Elazar, Yanai and Bengio, Yoshua},
  title = {Chain-of-Thought Is Not Explainability},
  publisher = {alphaXiv},
  year = {2025}
}

@book{krantz2002implicit,
  title={The implicit function theorem: history, theory, and applications},
  author={Krantz, Steven George and Parks, Harold R},
  year={2002},
  publisher={Springer Science \& Business Media}
}

@article{bricken2023monosemanticity,
   title={Towards Monosemanticity: Decomposing Language Models With Dictionary Learning},
   author={Bricken, Trenton and Templeton, Adly and Batson, Joshua and Chen, Brian and Jermyn, Adam and Conerly, Tom and Turner, Nick and Anil, Cem and Denison, Carson and Askell, Amanda and Lasenby, Robert and Wu, Yifan and Kravec, Shauna and Schiefer, Nicholas and Maxwell, Tim and Joseph, Nicholas and Hatfield-Dodds, Zac and Tamkin, Alex and Nguyen, Karina and McLean, Brayden and Burke, Josiah E and Hume, Tristan and Carter, Shan and Henighan, Tom and Olah, Christopher},
   year={2023},
   journal={Transformer Circuits Thread},
   note={https://transformer-circuits.pub/2023/monosemantic-features/index.html}
}

@article{topol2019high,
  title={High-performance medicine: the convergence of human and artificial intelligence},
  author={Topol, Eric J},
  journal={Nature medicine},
  volume={25},
  number={1},
  pages={44--56},
  year={2019},
  publisher={Nature Publishing Group US New York}
}

@article{ji2023survey,
  title={Survey of hallucination in natural language generation},
  author={Ji, Ziwei and Lee, Nayeon and Frieske, Rita and Yu, Tiezheng and Su, Dan and Xu, Yan and Ishii, Etsuko and Bang, Ye Jin and Madotto, Andrea and Fung, Pascale},
  journal={ACM computing surveys},
  volume={55},
  number={12},
  pages={1--38},
  year={2023},
  publisher={ACM New York, NY}
}

@inproceedings{oberst2019counterfactual,
  title={Counterfactual off-policy evaluation with gumbel-max structural causal models},
  author={Oberst, Michael and Sontag, David},
  booktitle={International Conference on Machine Learning},
  pages={4881--4890},
  year={2019},
  organization={PMLR}
}

@article{ghassemi2021false,
  title={The false hope of current approaches to explainable artificial intelligence in health care},
  author={Ghassemi, Marzyeh and Oakden-Rayner, Luke and Beam, Andrew L},
  journal={The lancet digital health},
  volume={3},
  number={11},
  pages={e745--e750},
  year={2021},
  publisher={Elsevier}
}

@article{futoma2020myth,
  title={The myth of generalisability in clinical research and machine learning in health care},
  author={Futoma, Joseph and Simons, Morgan and Panch, Trishan and Doshi-Velez, Finale and Celi, Leo Anthony},
  journal={The Lancet Digital Health},
  volume={2},
  number={9},
  pages={e489--e492},
  year={2020},
  publisher={Elsevier}
}

@article{turpin2023language,
  title={Language models don't always say what they think: Unfaithful explanations in chain-of-thought prompting},
  author={Turpin, Miles and Michael, Julian and Perez, Ethan and Bowman, Samuel},
  journal={Advances in Neural Information Processing Systems},
  volume={36},
  pages={74952--74965},
  year={2023}
}

@article{feldman2020neural,
  title={What neural networks memorize and why: Discovering the long tail via influence estimation},
  author={Feldman, Vitaly and Zhang, Chiyuan},
  journal={Advances in Neural Information Processing Systems},
  volume={33},
  pages={2881--2891},
  year={2020}
}

@article{wiegreffe2019attention,
  title={Attention is not not explanation},
  author={Wiegreffe, Sarah and Pinter, Yuval},
  journal={arXiv preprint arXiv:1908.04626},
  year={2019}
}

@article{jain2019attention,
  title={Attention is not explanation},
  author={Jain, Sarthak and Wallace, Byron C},
  journal={arXiv preprint arXiv:1902.10186},
  year={2019}
}

@inproceedings{sundararajan2017axiomatic,
  title={Axiomatic attribution for deep networks},
  author={Sundararajan, Mukund and Taly, Ankur and Yan, Qiqi},
  booktitle={International conference on machine learning},
  pages={3319--3328},
  year={2017},
  organization={PMLR}
}

@article{basu2020falsifiability,
  title={On the falsifiability and learnability of decision theories},
  author={Basu, Pathikrit and Echenique, Federico},
  journal={Theoretical Economics},
  volume={15},
  number={4},
  pages={1279--1305},
  year={2020},
  publisher={Wiley Online Library}
}

@article{tsimpoukelli2021multimodal,
  title={Multimodal few-shot learning with frozen language models},
  author={Tsimpoukelli, Maria and Menick, Jacob L and Cabi, Serkan and Eslami, SM and Vinyals, Oriol and Hill, Felix},
  journal={Advances in Neural Information Processing Systems},
  volume={34},
  pages={200--212},
  year={2021}
}

@techreport{cong2023sparse,
  title={Sparse modeling under grouped heterogeneity with an application to asset pricing},
  author={Cong, Lin William and Feng, Guanhao and He, Jingyu and Li, Junye},
  year={2023},
  institution={National Bureau of Economic Research}
}

@article{templeton2024scaling,
title={Scaling Monosemanticity: Extracting Interpretable Features from Claude 3 Sonnet},
author={Templeton, Adly and Conerly, Tom and Marcus, Jonathan and Lindsey, Jack and Bricken, Trenton and Chen, Brian and Pearce, Adam and Citro, Craig and Ameisen, Emmanuel and Jones, Andy and Cunningham, Hoagy and Turner, Nicholas L and McDougall, Callum and MacDiarmid, Monte and Freeman, C. Daniel and Sumers, Theodore R. and Rees, Edward and Batson, Joshua and Jermyn, Adam and Carter, Shan and Olah, Chris and Henighan, Tom},
year={2024},
journal={Transformer Circuits Thread},
url={https://transformer-circuits.pub/2024/scaling-monosemanticity/index.html}
}

@article{makhzani2013k,
  title={K-sparse autoencoders},
  author={Makhzani, Alireza and Frey, Brendan},
  journal={arXiv preprint arXiv:1312.5663},
  year={2013}
}

@inproceedings{surkov2025unpacking,
  title={Unpacking sdxl turbo: Interpreting text-to-image models with sparse autoencoders},
  author={Surkov, Viacheslav and Wendler, Chris and Terekhov, Mikhail and Deschenaux, Justin and West, Robert and Gulcehre, Caglar},
  booktitle={Mechanistic Interpretability for Vision at CVPR 2025 (Non-proceedings Track)},
  year={2025}
}

@article{abdulaal2024x,
  title={An x-ray is worth 15 features: Sparse autoencoders for interpretable radiology report generation},
  author={Abdulaal, Ahmed and Fry, Hugo and Monta{\~n}a-Brown, Nina and Ijishakin, Ayodeji and Gao, Jack and Hyland, Stephanie and Alexander, Daniel C and Castro, Daniel C},
  journal={arXiv preprint arXiv:2410.03334},
  year={2024}
}

@article{gao2024scaling,
  title={Scaling and evaluating sparse autoencoders},
  author={Gao, Leo and la Tour, Tom Dupr{\'e} and Tillman, Henk and Goh, Gabriel and Troll, Rajan and Radford, Alec and Sutskever, Ilya and Leike, Jan and Wu, Jeffrey},
  journal={arXiv preprint arXiv:2406.04093},
  year={2024}
}

@article{bussmann2024batchtopk,
  title={Batchtopk sparse autoencoders},
  author={Bussmann, Bart and Leask, Patrick and Nanda, Neel},
  journal={arXiv preprint arXiv:2412.06410},
  year={2024}
}

@article{meng_locating_2022,
  title = {Locating and Editing Factual Associations in GPT},
  author = {Meng, Kevin and Bau, David and Andonian, Alex and Belinkov, Yonatan},
  year = {2022},
  journal = {NeurIPS},
  eprint = {2202.05262},
  primaryclass = {cs},
  doi = {10.48550/arXiv.2202.05262},
  url = {http://arxiv.org/abs/2202.05262},
  urldate = {2023-08-27},
  archiveprefix = {arXiv}
}

@article{wang_interpretability_2023,
  title = {Interpretability in the Wild: a Circuit for Indirect Object Identification in GPT-2 small},
  shorttitle = {Interpretability in the Wild},
  author = {Wang, Kevin and Variengien, Alexandre and Conmy, Arthur and Shlegeris, Buck and Steinhardt, Jacob},
  year = {2023},
  journal = {ICLR},
  eprint = {2211.00593},
  primaryclass = {cs},
  doi = {10.48550/arXiv.2211.00593},
  url = {http://arxiv.org/abs/2211.00593},
  urldate = {2023-08-27},
  archiveprefix = {arXiv}
}

@article{mcdougall_copy_2023,
  title = {Copy Suppression: Comprehensively Understanding an Attention Head},
  shorttitle = {Copy Suppression},
  author = {McDougall, Callum and Conmy, Arthur and Rushing, Cody and McGrath, Thomas and Nanda, Neel},
  year = {2023},
  month = oct,
  journal = {CoRR},
  eprint = {2310.04625},
  primaryclass = {cs},
  doi = {10.48550/arXiv.2310.04625},
  url = {http://arxiv.org/abs/2310.04625},
  urldate = {2023-10-27},
  archiveprefix = {arXiv}
}

@article{baydin2017automatic,
  author  = {Atilim Gunes Baydin and Barak A. Pearlmutter and Alexey Andreyevich Radul and Jeffrey Mark Siskind},
  title   = {Automatic Differentiation in Machine Learning: a Survey},
  journal = {Journal of Machine Learning Research},
  year    = {2018},
  volume  = {18},
  number  = {153},
  pages   = {1--43},
  url     = {http://jmlr.org/papers/v18/17-468.html}
}

@article{cammarata_curve_2021,
  title = {Curve Circuits},
  author = {Cammarata, Nick and Goh, Gabriel and Carter, Shan and Voss, Chelsea and Schubert, Ludwig and Olah, Chris},
  year = {2021},
  journal = {Distill},
  url = {https://distill.pub/2020/circuits/curve-circuits/}
}

@article{olah_building_2018,
  title = {The Building Blocks of Interpretability},
  author = {Olah, Chris and Satyanarayan, Arvind and Johnson, Ian and Carter, Shan and Schubert, Ludwig and Ye, Katherine and Mordvintsev, Alexander},
  year = {2018},
  month = mar,
  journal = {Distill},
  url = {https://distill.pub/2018/building-blocks},
  urldate = {2023-05-31}
}

@article{cammarata_curve_2020,
  title = {Curve Detectors},
  author = {Cammarata, Nick and Goh, Gabriel and Carter, Shan and Schubert, Ludwig and Petrov, Michael and Olah, Chris},
  year = {2020},
  month = jun,
  journal = {Distill},
  url = {https://distill.pub/2020/circuits/curve-detectors},
  urldate = {2023-05-31}
}

@article{henighan_superposition_2023,
  title = {Superposition, Memorization, and Double Descent},
  author = {Henighan, Tom and Carter, Shan and Hume, Tristan and Elhage, Nelson and Lasenby, Robert and Fort, Stanislav and Schiefer, Nicholas and Olah, Christopher},
  year = {2023},
  journal = {Transformer Circuits Thread},
  url = {https://transformer-circuits.pub/2023/toy-double-descent/index.html}
}

@article{elhage_privileged_2023,
  title = {Privileged Bases in the Transformer Residual Stream},
  author = {Elhage, Nelson and Lasenby, Robert and Olah, Christopher},
  year = {2023},
  journal = {Transformer Circuits Thread},
  url = {https://transformer-circuits.pub/2023/privileged-basis/index.html}
}

@article{elhage_mathematical_2021,
  title = {A mathematical framework for transformer circuits},
  author = {Elhage, N and Nanda, N and Olsson, C and Henighan, T and Joseph, N and Mann, B and Askell, A and Bai, Y and Chen, A and Conerly, T and others},
  year = {2021},
  journal = {Transformer Circuits Thread},
  url = {https://transformer-circuits.pub/2021/framework/index.html}
}

@article{elhage_toy_2022,
  title = {Toy Models of Superposition},
  author = {Elhage, Nelson and Hume, Tristan and Olsson, Catherine and Schiefer, Nicholas and Henighan, Tom and Kravec, Shauna and {Hatfield-Dodds}, Zac and Lasenby, Robert and Drain, Dawn and Chen, Carol and others},
  year = {2022},
  journal = {Transformer Circuits Thread},
  url = {https://transformer-circuits.pub/2022/toy_model/index.html}
}

@article{elhage_softmax_2022,
  title = {Softmax Linear Units},
  author = {Elhage, Nelson and Hume, Tristan and Catherine, Olsson and Neel, Nanda and Henighan, Tom and Johnston, Scott and ElShowk, Sheer and Joseph, Nicholas and DasSarma, Nova and Mann, Ben and Hernandez, Danny and Askell, Amanda and Ndousse, Kamal and Drain, Dawn and Chen, Anna and Bai, Yuntao and Ganguli, Deep and Lovitt, Liane and {Hatfield-Dodds}, Zac and Kernion, Jackson and Conerly, Tom and Kravec, Shauna and Fort, Stanislav and Kadavath, Saurav and Jacobson, Josh and {Tran-Johnson}, Eli and Kaplan, Jared and Clark, Jack and Brown, Tom and McCandlish, Sam and Amodei, Dario and Olah, Christopher},
  year = {2022},
  journal = {Transformer Circuits Thread},
  url = {https://transformer-circuits.pub/2022/solu/index.html},
  urldate = {2023-07-31}
}

@article{olsson_incontext_2022,
  title = {In-context Learning and Induction Heads},
  author = {Olsson, Catherine and Elhage, Nelson and Nanda, Neel and Joseph, Nicholas and DasSarma, Nova and Henighan, Tom and Mann, Ben and Askell, Amanda and Bai, Yuntao and Chen, Anna and Conerly, Tom and Drain, Dawn and Ganguli, Deep and {Hatfield-Dodds}, Zac and Hernandez, Danny and Johnston, Scott and Jones, Andy and Kernion, Jackson and Lovitt, Liane and Ndousse, Kamal and Amodei, Dario and Brown, Tom and Clark, Jack and Kaplan, Jared and McCandlish, Sam and Olah, Chris},
  year = {2022},
  journal = {Transformer Circuits Thread},
  url = {https://transformer-circuits.pub/2022/in-context-learning-and-induction-heads/index.html}
}

@article{burns_discovering_2023,
  title = {Discovering Latent Knowledge in Language Models Without Supervision},
  author = {Burns, Collin and Ye, Haotian and Klein, Dan and Steinhardt, Jacob},
  year = {2023},
  journal = {ICLR},
  eprint = {2212.03827},
  primaryclass = {cs},
  url = {http://arxiv.org/abs/2212.03827},
  urldate = {2023-07-31},
  archiveprefix = {arXiv}
}

@article{gurnee_finding_2023,
  title = {Finding Neurons in a Haystack: Case Studies with Sparse Probing},
  shorttitle = {Finding Neurons in a Haystack},
  author = {Gurnee, Wes and Nanda, Neel and Pauly, Matthew and Harvey, Katherine and Troitskii, Dmitrii and Bertsimas, Dimitris},
  year = {2023},
  journal = {TMLR},
  publisher = {arXiv},
  url = {https://arxiv.org/abs/2305.01610},
  urldate = {2023-07-31},
  copyright = {Creative Commons Attribution 4.0 International}
}

@article{li_emergent_2023,
  title = {Emergent World Representations: Exploring a Sequence Model Trained on a Synthetic Task},
  shorttitle = {Emergent World Representations},
  author = {Li, Kenneth and Hopkins, Aspen K. and Bau, David and Vi{\'e}gas, Fernanda and Pfister, Hanspeter and Wattenberg, Martin},
  year = {2023},
  journal = {ICLR},
  publisher = {arXiv},
  url = {https://arxiv.org/abs/2210.13382},
  urldate = {2023-07-31},
  copyright = {Creative Commons Attribution 4.0 International}
}

@article{nanda_actually_2023,
  title = {Actually, Othello-GPT Has A Linear Emergent World Representation},
  author = {Nanda, Neel},
  year = {2023},
  month = mar,
  journal = {Neel Nanda's Blog},
  url = {https://neelnanda.io/mechanistic-interpretability/othello}
}

@article{black_interpreting_2022,
  title = {Interpreting Neural Networks through the Polytope Lens},
  author = {Black, Sid and Sharkey, Lee and Grinsztajn, Leo and Winsor, Eric and Braun, Dan and Merizian, Jacob and Parker, Kip and Guevara, Carlos Ram{\'o}n and Millidge, Beren and Alfour, Gabriel and Leahy, Connor},
  year = {2022},
  month = nov,
  journal = {CoRR},
  publisher = {arXiv},
  url = {https://arxiv.org/abs/2211.12312},
  urldate = {2023-07-31},
  copyright = {arXiv.org perpetual, non-exclusive license}
}

@article{scherlis_polysemanticity_2023,
  title = {Polysemanticity and Capacity in Neural Networks},
  author = {Scherlis, Adam and Sachan, Kshitij and Jermyn, Adam S. and Benton, Joe and Shlegeris, Buck},
  year = {2023},
  month = jul,
  journal = {CoRR},
  eprint = {2210.01892},
  primaryclass = {cs},
  doi = {10.48550/arXiv.2210.01892},
  url = {http://arxiv.org/abs/2210.01892},
  urldate = {2023-09-18},
  archiveprefix = {arXiv}
}

@article{sharkey_circumventing_2022,
  title = {Circumventing interpretability: How to defeat mind-readers},
  shorttitle = {Circumventing interpretability},
  author = {Sharkey, Lee},
  year = {2022},
  month = dec,
  journal = {CoRR},
  publisher = {arXiv},
  doi = {10.48550/ARXIV.2212.11415},
  url = {https://arxiv.org/abs/2212.11415},
  urldate = {2023-09-18},
  copyright = {Creative Commons Attribution 4.0 International}
}

@article{geiger_causal_2021,
  title = {Causal Abstractions of Neural Networks},
  author = {Geiger, Atticus and Lu, Hanson and Icard, Thomas and Potts, Christopher},
  year = {2021},
  journal = {NeurIPS},
  volume = {34},
  pages = {9574--9586},
  url = {https://proceedings.neurips.cc/paper/2021/hash/4f5c422f4d49a5a807eda27434231040-Abstract.html},
  urldate = {2023-08-29}
}

@article{zou_representation_2023,
  title = {Representation Engineering: A Top-Down Approach to AI Transparency},
  shorttitle = {Representation Engineering},
  author = {Zou, Andy and Phan, Long and Chen, Sarah and Campbell, James and Guo, Phillip and Ren, Richard and Pan, Alexander and Yin, Xuwang and Mazeika, Mantas and Dombrowski, Ann-Kathrin and Goel, Shashwat and Li, Nathaniel and Byun, Michael J. and Wang, Zifan and Mallen, Alex and Basart, Steven and Koyejo, Sanmi and Song, Dawn and Fredrikson, Matt and Kolter, J. Zico and Hendrycks, Dan},
  year = {2023},
  month = oct,
  journal = {CoRR},
  eprint = {2310.01405},
  primaryclass = {cs},
  doi = {10.48550/arXiv.2310.01405},
  url = {http://arxiv.org/abs/2310.01405},
  urldate = {2023-11-08},
  archiveprefix = {arXiv}
}

@article{hendel_incontext_2023,
  title = {In-Context Learning Creates Task Vectors},
  author = {Hendel, Roee and Geva, Mor and Globerson, Amir},
  year = {2023},
  month = oct,
  journal = {EMNLP},
  eprint = {2310.15916},
  primaryclass = {cs},
  doi = {10.48550/arXiv.2310.15916},
  url = {http://arxiv.org/abs/2310.15916},
  urldate = {2023-11-08},
  archiveprefix = {arXiv}
}

@article{omahony_disentangling_2023,
  title = {Disentangling Neuron Representations with Concept Vectors},
  author = {O'Mahony, Laura and Andrearczyk, Vincent and Muller, Henning and Graziani, Mara},
  year = {2023},
  month = apr,
  journal = {CVPR Workshops},
  eprint = {2304.09707},
  primaryclass = {cs},
  doi = {10.48550/arXiv.2304.09707},
  url = {http://arxiv.org/abs/2304.09707},
  urldate = {2023-09-18},
  archiveprefix = {arXiv}
}

@article{deng_measuring_2023,
  title = {Measuring Feature Sparsity in Language Models},
  author = {Deng, Mingyang and Tao, Lucas and Benton, Joe},
  year = {2023},
  journal = {CoRR},
  publisher = {arXiv},
  doi = {10.48550/ARXIV.2310.07837},
  url = {https://arxiv.org/abs/2310.07837},
  urldate = {2023-10-26},
  copyright = {arXiv.org perpetual, non-exclusive license}
}

@article{janus_simulators_2022,
  title = {Simulators},
  author = {{janus}},
  year = {2022},
  month = sep,
  journal = {LessWrong},
  url = {https://www.lesswrong.com/posts/vJFdjigzmcXMhNTsx/simulators},
  urldate = {2023-05-15},
  language = {en}
}

@article{belinkov_probing_2022,
  title = {Probing Classifiers: Promises, Shortcomings, and Advances},
  shorttitle = {Probing Classifiers},
  author = {Belinkov, Yonatan},
  year = {2022},
  month = mar,
  journal = {Computational Linguistics},
  volume = {48},
  number = {1},
  pages = {207--219},
  publisher = {MIT Press},
  address = {Cambridge, MA},
  doi = {10.1162/coli_a_00422},
  url = {https://aclanthology.org/2022.cl-1.7},
  urldate = {2023-11-10}
}

@article{hernandez_linearity_2023,
  title = {Linearity of Relation Decoding in Transformer Language Models},
  author = {Hernandez, Evan and Sharma, Arnab Sen and Haklay, Tal and Meng, Kevin and Wattenberg, Martin and Andreas, Jacob and Belinkov, Yonatan and Bau, David},
  year = {2023},
  month = aug,
  journal = {CoRR},
  eprint = {2308.09124},
  primaryclass = {cs},
  doi = {10.48550/arXiv.2308.09124},
  url = {http://arxiv.org/abs/2308.09124},
  urldate = {2023-11-16},
  archiveprefix = {arXiv}
}

@article{ribeiro_why_2016,
  title = {"Why Should I Trust You?": Explaining the Predictions of Any Classifier},
  shorttitle = {"Why Should I Trust You?},
  author = {Ribeiro, Marco Tulio and Singh, Sameer and Guestrin, Carlos},
  year = {2016},
  month = aug,
  journal = {NAACL},
  eprint = {1602.04938},
  primaryclass = {cs, stat},
  doi = {10.48550/arXiv.1602.04938},
  url = {http://arxiv.org/abs/1602.04938},
  urldate = {2023-11-16},
  archiveprefix = {arXiv}
}

@article{ivanitskiy_structured_2023,
  title = {Structured World Representations in Maze-Solving Transformers},
  author = {Ivanitskiy, M. and Spies, Alexander F. and Rauker, Tilman and Corlouer, Guillaume and Mathwin, Chris and Quirke, Lucia and Rager, Can and Shah, Rusheb and Valentine, Dan and Behn, Cecilia Diniz and Inoue, Katsumi and Fung, Samy Wu},
  year = {2023},
  month = dec,
  journal = {CoRR},
  url = {https://arxiv.org/abs/2312.02566},
  urldate = {2023-12-11}
}

@article{todd_function_2023,
  title = {Function Vectors in Large Language Models},
  author = {Todd, Eric and Li, Millicent L. and Sharma, Arnab Sen and Mueller, Aaron and Wallace, Byron C. and Bau, David},
  year = {2023},
  journal = {CoRR},
  publisher = {arXiv},
  doi = {10.48550/ARXIV.2310.15213},
  url = {https://arxiv.org/abs/2310.15213},
  urldate = {2023-12-11},
  copyright = {arXiv.org perpetual, non-exclusive license}
}

@article{chan_natural_2023,
  title = {Natural Abstractions: Key claims, Theorems, and Critiques},
  shorttitle = {Natural Abstractions},
  author = {Chan, Lawrence and Lang, Leon and Jenner, Erik},
  year = {2023},
  month = mar,
  journal = {AI Alignment Forum},
  url = {https://www.alignmentforum.org/posts/gvzW46Z3BsaZsLc25/natural-abstractions-key-claims-theorems-and-critiques-1},
  urldate = {2024-01-17},
  language = {en}
}

@article{kornblith_similarity_2019,
  title = {Similarity of Neural Network Representations Revisited},
  author = {Kornblith, Simon and Norouzi, Mohammad and Lee, Honglak and Hinton, Geoffrey},
  year = {2019},
  month = jul,
  journal = {ICML},
  eprint = {1905.00414},
  primaryclass = {cs, q-bio, stat},
  doi = {10.48550/arXiv.1905.00414},
  url = {http://arxiv.org/abs/1905.00414},
  urldate = {2024-01-18},
  archiveprefix = {arXiv}
}

@article{warstadt_blimp_2020,
  title = {BLiMP: The Benchmark of Linguistic Minimal Pairs for English},
  shorttitle = {BLiMP},
  author = {Warstadt, Alex and Parrish, Alicia and Liu, Haokun and Mohananey, Anhad and Peng, Wei and Wang, Sheng-Fu and Bowman, Samuel R.},
  editor = {Johnson, Mark and Roark, Brian and Nenkova, Ani},
  year = {2020},
  journal = {Transactions of the Association for Computational Linguistics},
  volume = {8},
  pages = {377--392},
  publisher = {MIT Press},
  address = {Cambridge, MA},
  doi = {10.1162/tacl_a_00321},
  url = {https://aclanthology.org/2020.tacl-1.25},
  urldate = {2024-01-24}
}

@article{bach_pixelwise_2015,
  title = {On Pixel-Wise Explanations for Non-Linear Classifier Decisions by Layer-Wise Relevance Propagation},
  author = {Bach, Sebastian and Binder, Alexander and Montavon, Gr{\'e}goire and Klauschen, Frederick and M{\"u}ller, Klaus-Robert and Samek, Wojciech},
  year = {2015},
  month = jul,
  journal = {PLOS ONE},
  volume = {10},
  number = {7},
  pages = {e0130140},
  publisher = {Public Library of Science},
  issn = {1932-6203},
  doi = {10.1371/journal.pone.0130140},
  url = {https://journals.plos.org/plosone/article?id=10.1371/journal.pone.0130140},
  urldate = {2024-01-24},
  language = {en}
}

@article{covert_explaining_2021,
  title = {Explaining by removing: a unified framework for model explanation},
  shorttitle = {Explaining by removing},
  author = {Covert, Ian C. and Lundberg, Scott and Lee, Su-In},
  year = {2021},
  month = jan,
  journal = {J. Mach. Learn. Res.},
  volume = {22},
  number = {1},
  pages = {209:9477--209:9566},
  issn = {1532-4435},
  url = {https://arxiv.org/abs/2011.14878}
}

@article{shrikumar_learning_2017,
  title = {Learning Important Features Through Propagating Activation Differences},
  author = {Shrikumar, Avanti and Greenside, Peyton and Kundaje, Anshul},
  year = {2017},
  journal = {ICML},
  eprint = {1704.02685},
  primaryclass = {cs},
  doi = {10.48550/arXiv.1704.02685},
  url = {http://arxiv.org/abs/1704.02685},
  urldate = {2024-01-24},
  archiveprefix = {arXiv}
}

@article{selvaraju_gradcam_2016,
  title = {Grad-CAM: Why did you say that? Visual Explanations from Deep Networks via Gradient-based Localization},
  shorttitle = {Grad-CAM},
  author = {Selvaraju, Ramprasaath R. and Das, Abhishek and Vedantam, Ramakrishna and Cogswell, Michael and Parikh, Devi and Batra, Dhruv},
  year = {2016},
  journal = {ICCV},
  url = {https://arxiv.org/abs/1610.02391},
  urldate = {2024-01-24}
}

@article{sundararajan_axiomatic_2017,
  title = {Axiomatic Attribution for Deep Networks},
  author = {Sundararajan, Mukund and Taly, Ankur and Yan, Qiqi},
  year = {2017},
  month = jun,
  journal = {ICML},
  eprint = {1703.01365},
  primaryclass = {cs},
  doi = {10.48550/arXiv.1703.01365},
  url = {http://arxiv.org/abs/1703.01365},
  urldate = {2024-01-24},
  archiveprefix = {arXiv}
}

@article{casalicchio_visualizing_2018,
  title = {Visualizing the Feature Importance for Black Box Models},
  author = {Casalicchio, Giuseppe and Molnar, Christoph and Bischl, Bernd},
  year = {2018},
  journal = {ECML PKDD},
  volume = {11051},
  eprint = {1804.06620},
  primaryclass = {cs, stat},
  pages = {655--670},
  doi = {10.1007/978-3-030-10925-7_40},
  url = {http://arxiv.org/abs/1804.06620},
  urldate = {2024-01-24},
  archiveprefix = {arXiv}
}

@article{smilkov_smoothgrad_2017,
  title = {SmoothGrad: removing noise by adding noise},
  shorttitle = {SmoothGrad},
  author = {Smilkov, Daniel and Thorat, Nikhil and Kim, Been and Vi{\'e}gas, Fernanda and Wattenberg, Martin},
  year = {2017},
  month = jun,
  journal = {CoRR},
  eprint = {1706.03825},
  primaryclass = {cs, stat},
  doi = {10.48550/arXiv.1706.03825},
  url = {http://arxiv.org/abs/1706.03825},
  urldate = {2024-01-24},
  archiveprefix = {arXiv}
}

@article{chanin_identifying_2023,
  title = {Identifying Linear Relational Concepts in Large Language Models},
  author = {Chanin, David and Hunter, Anthony and Camburu, Oana-Maria},
  year = {2023},
  journal = {CoRR},
  publisher = {arXiv},
  doi = {10.48550/ARXIV.2311.08968},
  url = {https://arxiv.org/abs/2311.08968},
  urldate = {2024-02-11},
  copyright = {arXiv.org perpetual, non-exclusive license}
}

@article{nanda_200algorithm_2023,
  title = {200 COP in MI: Interpreting Algorithmic Problems},
  shorttitle = {200 COP in MI},
  author = {Nanda, Neel},
  year = {2022},
  journal = {Neel Nanda's Blog},
  url = {https://www.lesswrong.com/posts/ejtFsvyhRkMofKAFy/200-cop-in-mi-interpreting-algorithmic-problems},
  urldate = {2024-03-01},
  language = {en}
}

@article{bilodeau_impossibility_2024,
  title = {Impossibility Theorems for Feature Attribution},
  author = {Bilodeau, Blair and Jaques, Natasha and Koh, Pang Wei and Kim, Been},
  year = {2024},
  month = jan,
  journal = {Proc. Natl. Acad. Sci. U.S.A.},
  volume = {121},
  number = {2},
  eprint = {2212.11870},
  primaryclass = {cs},
  pages = {e2304406120},
  issn = {0027-8424, 1091-6490},
  doi = {10.1073/pnas.2304406120},
  url = {http://arxiv.org/abs/2212.11870},
  urldate = {2024-03-05},
  archiveprefix = {arXiv}
}

@article{bansal_revisiting_2021,
  title = {Revisiting Model Stitching to Compare Neural Representations},
  author = {Bansal, Yamini and Nakkiran, Preetum and Barak, Boaz},
  year = {2021},
  month = jun,
  journal = {CoRR},
  eprint = {2106.07682},
  primaryclass = {cs, stat},
  doi = {10.48550/arXiv.2106.07682},
  url = {http://arxiv.org/abs/2106.07682},
  urldate = {2024-03-19},
  archiveprefix = {arXiv}
}

@article{marchetti_harmonics_2023,
  title = {Harmonics of Learning: Universal Fourier Features Emerge in Invariant Networks},
  shorttitle = {Harmonics of Learning},
  author = {Marchetti, Giovanni Luca and Hillar, Christopher and Kragic, Danica and Sanborn, Sophia},
  year = {2023},
  month = dec,
  journal = {CoRR},
  eprint = {2312.08550},
  primaryclass = {cs, eess},
  doi = {10.48550/arXiv.2312.08550},
  url = {http://arxiv.org/abs/2312.08550},
  urldate = {2024-05-07},
  archiveprefix = {arXiv}
}

@article{engels_not_2024,
  title = {Not All Language Model Features Are Linear},
  author = {Engels, Joshua and Liao, Isaac and Michaud, Eric J. and Gurnee, Wes and Tegmark, Max},
  year = {2024},
  month = may,
  journal = {CoRR},
  eprint = {2405.14860},
  primaryclass = {cs},
  doi = {10.48550/arXiv.2405.14860},
  url = {http://arxiv.org/abs/2405.14860},
  urldate = {2024-06-06},
  archiveprefix = {arXiv}
}

@article{templeton_scaling_2024,
  title = {Scaling Monosemanticity: Extracting Interpretable Features from Claude 3 Sonnet},
  author = {Templeton, Adly and Conerly, Tom and Marcus, Jonathan and Lindsey, Jack and Bricken, Trenton and Chen, Brian},
  year = {2024},
  journal = {Transformer Circuits Thread},
  url = {https://transformer-circuits.pub/2024/scaling-monosemanticity/index.html}
}

@article{rajamanoharan_improving_2024,
  title = {Improving Dictionary Learning with Gated Sparse Autoencoders},
  author = {Rajamanoharan, Senthooran and Conmy, Arthur and Smith, Lewis and Lieberum, Tom and Varma, Vikrant and Kram{\'a}r, J{\'a}nos and Shah, Rohin and Nanda, Neel},
  year = {2024},
  month = apr,
  journal = {CoRR},
  eprint = {2404.16014},
  primaryclass = {cs},
  doi = {10.48550/arXiv.2404.16014},
  url = {http://arxiv.org/abs/2404.16014},
  urldate = {2024-06-10},
  archiveprefix = {arXiv}
}

@article{shanahan_role_2023,
  title = {Role play with large language models},
  author = {Shanahan, Murray and McDonell, Kyle and Reynolds, Laria},
  year = {2023},
  month = nov,
  journal = {Nature},
  volume = {623},
  number = {7987},
  pages = {493--498},
  publisher = {Nature Publishing Group},
  issn = {1476-4687},
  doi = {10.1038/s41586-023-06647-8},
  url = {https://www.nature.com/articles/s41586-023-06647-8},
  urldate = {2024-07-11},
  copyright = {2023 Springer Nature Limited},
  language = {en}
}

@article{karvonen_emergent_2024,
  title = {Emergent World Models and Latent Variable Estimation in Chess-Playing Language Models},
  author = {Karvonen, Adam},
  year = {2024},
  month = jul,
  journal = {COLM},
  eprint = {2403.15498},
  primaryclass = {cs},
  doi = {10.48550/arXiv.2403.15498},
  url = {http://arxiv.org/abs/2403.15498},
  urldate = {2024-08-06},
  archiveprefix = {arXiv}
}

@article{arditi_refusal_2024,
  title = {Refusal in Language Models Is Mediated by a Single Direction},
  author = {Arditi, Andy and Obeso, Oscar and Syed, Aaquib and Paleka, Daniel and Panickssery, Nina and Gurnee, Wes and Nanda, Neel},
  year = {2024},
  journal = {CoRR},
  publisher = {arXiv},
  doi = {10.48550/ARXIV.2406.11717},
  url = {https://arxiv.org/abs/2406.11717},
  urldate = {2024-08-08},
  copyright = {arXiv.org perpetual, non-exclusive license}
}

@article{park_geometry_2024,
  title = {The Geometry of Categorical and Hierarchical Concepts in Large Language Models},
  author = {Park, Kiho and Choe, Yo Joong and Jiang, Yibo and Veitch, Victor},
  year = {2024},
  month = jun,
  journal = {ICML MI Workshop (Oral)},
  url = {https://openreview.net/forum?id=KXuYjuBzKo},
  urldate = {2024-08-16},
  language = {en}
}

@article{tigges_language_2024,
  title = {Language Models Linearly Represent Sentiment},
  author = {Tigges, Curt and Hollinsworth, Oskar John and Geiger, Atticus and Nanda, Neel},
  year = {2024},
  month = jun,
  journal = {ICML MI Workshop},
  url = {https://openreview.net/forum?id=Xsf6dOOMMc},
  urldate = {2024-08-16},
  language = {en}
}

@article{wang2024disentangled,
  title={Disentangled Representation Learning},
  author={Wang, Xin and Guo, Hong and Jha, Sumit and Gao, Ruishan},
  journal={arXiv preprint arXiv:2211.11695},
  year={2024}
}

@inproceedings{
dunefsky2024transcoders,
title={Transcoders find interpretable {LLM} feature circuits},
author={Jacob Dunefsky and Philippe Chlenski and Neel Nanda},
booktitle={The Thirty-eighth Annual Conference on Neural Information Processing Systems},
year={2024},
url={https://openreview.net/forum?id=J6zHcScAo0}
}

@article{han2023medalpaca,
  title={MedAlpaca--An Open-Source Collection of Medical Conversational AI Models and Training Data},
  author={Han, Tianyu and Adams, Lisa C and Papaioannou, Jens-Michalis and Grundmann, Paul and Oberhauser, Tom and L{\"o}ser, Alexander and Truhn, Daniel and Bressem, Keno K},
  journal={arXiv preprint arXiv:2304.08247},
  year={2023}
}

@inproceedings{mihaylov2018openbookqa,
  title     = {Can a Suit of Armor Conduct Electricity? 
               A New Dataset for Open Book Question Answering},
  author    = {Mihaylov, Todor and Clark, Peter 
               and Khot, Tushar and Sabharwal, Ashish},
  booktitle = {Proceedings of the 2018 Conference on Empirical Methods in Natural Language Processing (EMNLP)},
  pages     = {2381--2391},
  year      = {2018},
  publisher = {Association for Computational Linguistics},
  doi       = {10.18653/v1/D18-1260}
}

@inproceedings{talmor2019commonsenseqa,
  title     = {CommonsenseQA: A Question Answering Challenge 
               Targeting Commonsense Knowledge},
  author    = {Talmor, Alon and Herzig, Jonathan 
               and Lourie, Nicholas and Berant, Jonathan},
  booktitle = {Proceedings of the 2019 Conference of the North 
               American Chapter of the Association for 
               Computational Linguistics (NAACL)},
  pages     = {4149--4158},
  year      = {2019}
}

@incollection{lundberg2017unified,
  title = {A Unified Approach to Interpreting Model Predictions},
  author = {Lundberg, Scott M and Lee, Su-In},
  booktitle = {Advances in Neural Information Processing Systems 30},
  pages = {4765--4774},
  year = {2017},
  publisher = {Curran Associates, Inc.}
}

@article{fel2025archetypal,
  title={Archetypal SAE: Adaptive and Stable Dictionary Learning for Concept Extraction in Large Vision Models},
  author={Fel, Thomas and others},
  journal={arXiv preprint},
  year={2025}
}
